\documentclass{article}

\usepackage[final]{corl_2026} 
\usepackage{amsmath}
\usepackage{amssymb}
\usepackage{booktabs}
\usepackage{tabularx}
\usepackage{xspace}
\usepackage{xcolor}
\usepackage{graphicx} 
\usepackage{enumitem}
\usepackage{multirow}
\usepackage{makecell}
\usepackage{float}
\usepackage[dvipsnames]{xcolor}
\newcommand{\ours}{Foresight\xspace}

\title{\ours: Failure Detection for Long-Horizon Robotic Manipulation with Action-Conditioned World Model Latents}

\author{%
\normalfont
Haoran Zhang$^{1,*}$,
Yifu Lu$^{2,*}$,
Boyang Wang$^{3}$,
Xuhui Kang$^{3}$, \\[0.2em]
Yen-Ling Kuo$^{3}$,
Zezhou Cheng$^{3}$,
Mengdi Wang$^{2}$,
Odest Chadwicke Jenkins$^{1,\dagger}$
\\[0.8em]
$^{1}$University of Michigan \quad
$^{2}$Princeton University \quad
$^{3}$University of Virginia
\\[0.6em]
{\small
$^*$Equal contribution. \quad
$^\dagger$Corresponding author.
}
\\[0.9em]
{\small
\textbf{Project Page: }
\href{https://haoranzhangumich.github.io/Forsight_web}
{\textcolor{blue}{Foresight.github.io}}
}
}

\begin{document}
\maketitle

\begin{abstract}
Long-horizon tasks are common in real-world robotic deployments, yet failure detection for such tasks remains underexplored. Detecting failures in long-horizon robotic tasks is particularly challenging because failure onset is often ambiguous and dense temporal annotations are typically unavailable. We present \ours, a failure detection framework that monitors manipulation trajectories using latent representations from an action-conditioned world model. \ours is trained using only final task-level success or failure labels. By leveraging predictive world-model embeddings, our method provides a unified framework for failure detection across different policies. We further use functional conformal prediction (FCP) to calibrate detection thresholds adaptively. We evaluate \ours with state-of-the-art vision-language-action policies in simulation on LIBERO-Long, ManiSkill-Long, and BEHAVIOR-1K, compare it against state-of-the-art failure detection methods, and validate it on real robots with three long-horizon tasks on a ReactorX-200 arm and one task on a Franka arm. Our results suggest that action-conditioned world-model embeddings provide a scalable representation for reliable failure monitoring in long-horizon manipulation. 
\end{abstract}



\section{Introduction}

\label{sec:introduction}
Robots operating over long horizons must recognize not only when a task has failed, but also when an ongoing execution has drifted toward failure. We study failure detection: given the observations and actions available up to time $t$, a detector assigns a failure score to the current rollout. Prior work has estimated this score using policy uncertainty~\citep{xu2025can}, policy-internal representations~\citep{gu2025safe,yeh2024aed}, vision-language judgments of visible mistakes~\citep{duan2024aha}, embedding distribution~\citep{he2024rediffuser}, or world-model latents~\citep{bardes2024revisiting, ho2026worldmodel}. These methods show that failures can often be detected before a terminal outcome is observed. However, most focus on short-horizon settings, isolated visual anomalies, or policy-specific confidence signals, leaving failure detection in long-horizon robotic tasks underexplored.

Long-horizon failure detection is challenging because the meaning of a visual state depends on the action history and task stage. The same object resting on a table may be expected before a grasp, evidence of a missed grasp after a lift command, or correct after a placement action. In multi-stage tasks lasting hundreds or thousands of steps, small deviations can accumulate and only later become irreversible. Effective detectors must therefore look beyond whether the current image appears unusual; they must judge whether the observed trajectory remains consistent with the progress implied by the robot's actions.

This connection motivates us to use an action-conditioned (AC) world model~\citep{assran2025vjepa2} as the backbone for long-horizon failure detection. 
Latent representations from action-conditioned world models compactly encode task-relevant state cues, including spatial relationships, motion patterns, interaction dynamics, and action-conditioned scene changes. By condensing task-relevant state cues into a small set of informative tokens, these representations are well-suited for monitoring long-horizon tasks.

We introduce \ours, a policy-interface-agnostic failure detector for long-horizon robotic tasks that leverages latent representations from an action-conditioned world model.
We freeze the visual encoder of the pretrained world model~\citep{assran2025vjepa2} and then attach and train an action-conditioned predictor from scratch.
The resulting action-conditioned world model produces predicted latent features for rollouts, which are then passed to the downstream failure detector, a simple yet effective causal Transformer~\citep{transformer}. 
Finally, we calibrate the detector with conformal prediction on held-out successful rollouts, yielding time-varying thresholds~\citep{vovk2005algorithmic, diquigiovanni2025importance}. 
\ours does not require policy logits, hidden states, token probabilities, or access to a policy-specific uncertainty head; it only uses the rollout interface of visual observations and the corresponding action chunks. As a result, the same framework can be applied to different vision-language-action (VLA) and visuomotor policies once the dataset-specific AC predictor and detector are trained.

To fully demonstrate the effectiveness of \ours, we comprehensively evaluate on challenging long-horizon simulation benchmarks and real-robot rollouts. 
In simulation, we adopt LIBERO-Long~\citep{liu2023libero}, ManiSkill-Long~\citep{taomaniskill3}, and BEHAVIOR-1K~\citep{li2024behavior1k}, covering tabletop manipulation and mobile household tasks. 
These settings include horizons from hundreds of steps to BEHAVIOR-1K rollouts averaging more than 8{,}000 steps. We collect rollouts from multiple policy families, including OpenVLA~\citep{kim24openvla}, SmolVLA~\citep{shukor2025smolvla}, $\pi_0$-FAST~\citep{black2024pi0,pertsch2025fast}, and $\pi_{0.5}$ policy~\citep{black2025pi05}. We also test real ReactorX and Franka robot rollouts with ACT~\citep{zhao2023learning}, $\pi_{0.5}$, SmolVLA, and GR00T N1.5 policies~\citep{gr00tn1_2025}. 
Across these benchmarks, we compare against multiple state-of-the-art baselines.

Our main contributions are:
\begin{itemize}[leftmargin=*, itemsep=0.1em, topsep=0.1em, parsep=0pt]

\item We propose \ours, a failure detection framework for long-horizon robotic manipulation that feeds latent representations from an action-conditioned world model consisting of a frozen visual encoder and trained AC predictor as inputs to a causal transformer failure detector.

\item We show that action-conditioned world model embeddings enable failure detection with supervision from only final task success/failure labels across different vision-language-action policies, and we incorporate functional conformal prediction to adaptively calibrate detection thresholds for reliable long-horizon failure detection.

\item We provide a comprehensive evaluation of long-horizon failure detection across diverse manipulation tasks, policies, robotic embodiments, simulation benchmarks, and real-world experiments, demonstrating the effectiveness of \ours against state-of-the-art failure detection methods.

\end{itemize}

\section{Related Work}
\label{sec:related_works}

\subsection{Failure Detection}
Failure detection aims to identify unsuccessful robot executions from partial or complete rollout observations. 
Prior work has explored different monitoring signals. 
Some works~\citep{AgiaSinhaEtAl2024, duan2024aha, pacaud2026scalingcrossenvironmentfailurereasoning, yi2026criticlooptrisystemvla,zhou2025codeasmonitorconstraintawarevisualprogramming, ifailsense2026} frame failure detection as a vision-language reasoning problem, using a vision-language model to detect manipulation failures and provide natural-language explanations. 
ReDiffuser~\citep{he2024rediffuser} learns a confidence function based on Random Network Distillation (RND) to measure the reliability of sampled decisions. 
FAIL-Detect~\citep{xu2025can} formulates failure detection for imitation-learning policies as sequential out-of-distribution detection, extracting scalar failure scores from policy observations and predicted actions, and calibrating time-varying thresholds with conformal prediction. 
SAFE~\citep{gu2025safe} instead studies multitask failure detection for vision-language-action policies, training lightweight detectors on policy-internal representations to predict per-step failure scores from trajectory-level outcome labels. 
More recently, some works explore using world models as a signal~\citep{ward2026foundationalworldmodelsaccurately,ho2026worldmodel,liu2024multitaskinteractiverobotfleet}. For instance, Gauge~\citep{ho2026worldmodel} uses compressed video world-model latents~\citep{agarwal2025cosmos} with conformal prediction thresholds to classify executions as success, known failure, or out-of-distribution anomaly.

\subsection{Foundational World Models}
Foundation world models aim to learn general-purpose representations or simulators of physical dynamics from large-scale video data. 
This has motivated a growing line of video-based world models~\cite{du2023learning, wang2025language} that repurpose generative video prediction for simulating future observations.
Cosmos World Foundation Models~\citep{agarwal2025cosmos} introduce a generative world model platform for physical AI, with Cosmos-Predict models supporting future video or world-state prediction from text, image, or video conditions. 
Cosmos-Predict2.5~\citep{nvidia2025worldsimulationvideofoundation} extends this line with video foundation models for world simulation, supporting conditional generation from text, images, and videos. 
In parallel, joint-embedding predictive architectures learn world representations without reconstructing pixels. 
V-JEPA~\citep{bardes2024revisiting} predicts masked video regions in latent space, and V-JEPA 2~\citep{assran2025vjepa2} scales this idea with internet scale video pretraining and post-trains an action-conditioned variant, V-JEPA 2-AC, on robot trajectories for physical prediction and planning. 
Because V-JEPA 2-AC predicts future latent states conditioned on robot actions, its representations can capture whether the observed execution is consistent with the policy's intended behavior. We therefore use these action-conditioned world-model features as signals for detecting failures in long-horizon manipulation tasks.

\section{Problem Formulation}

We study failure detection for robot policies in long-horizon manipulation tasks. In this work, we define long-horizon tasks as those requiring multiple subgoals, typically involving multiple symbolic manipulation actions such as pick, place, open, and close. At timestep $t$, the robot receives an image observation $I_t$. Let $c_t$ denote the observation context available at timestep $t$, which may consist of the current image $I_t$ or a short history of recent images.

A robot policy $\pi$ maps the observation context $c_t$ to a predicted action chunk
\begin{equation}
A_t = \left(a_{t|t}, a_{t+1|t}, \dots, a_{t+H-1|t}\right),
\end{equation}
where $H$ denotes the prediction horizon and $a_{t+k|t}$ denotes the action predicted for timestep $t+k$ at replanning timestep $t$. The robot executes the first $H' \leq H$ actions from this chunk before replanning.

Each completed rollout is annotated with a trajectory-level binary outcome label
\begin{equation}
y =
\begin{cases}
1, & \text{if the robot fails to complete the task}, \\
0, & \text{if the robot successfully completes the task}.
\end{cases}
\end{equation}
We assume access only to trajectory-level success or failure labels, without annotations of the precise timestep at which a failure occurs.

At execution time, our goal is to predict whether the ongoing rollout will eventually fail using only information available before executing the next action chunk. Given the current observation context $c_t$ and the policy-predicted action chunk $A_t$, we formulate failure detection as learning a scoring function
\begin{equation}
D_\theta \colon \{(c_i, A_i)\}_{i=1}^{t} = s_t
\end{equation}
where $s_t \in [0,1]$ is the predicted failure score at timestep $t$.

A failure alarm is triggered when the score exceeds a time-varying decision threshold $\delta_t$:
\begin{equation}
\hat{y}_t =
\begin{cases}
1, & \text{if } s_t \geq \delta_t, \\
0, & \text{otherwise}.
\end{cases}
\end{equation}
The objective is to detect impending or ongoing failures during execution, while using only the current observation context and the policy's next predicted action chunk.

 \section{Methodology}

\begin{figure}[t]
\centering
\vspace{-0.4cm}
\includegraphics[width=1.0\linewidth]{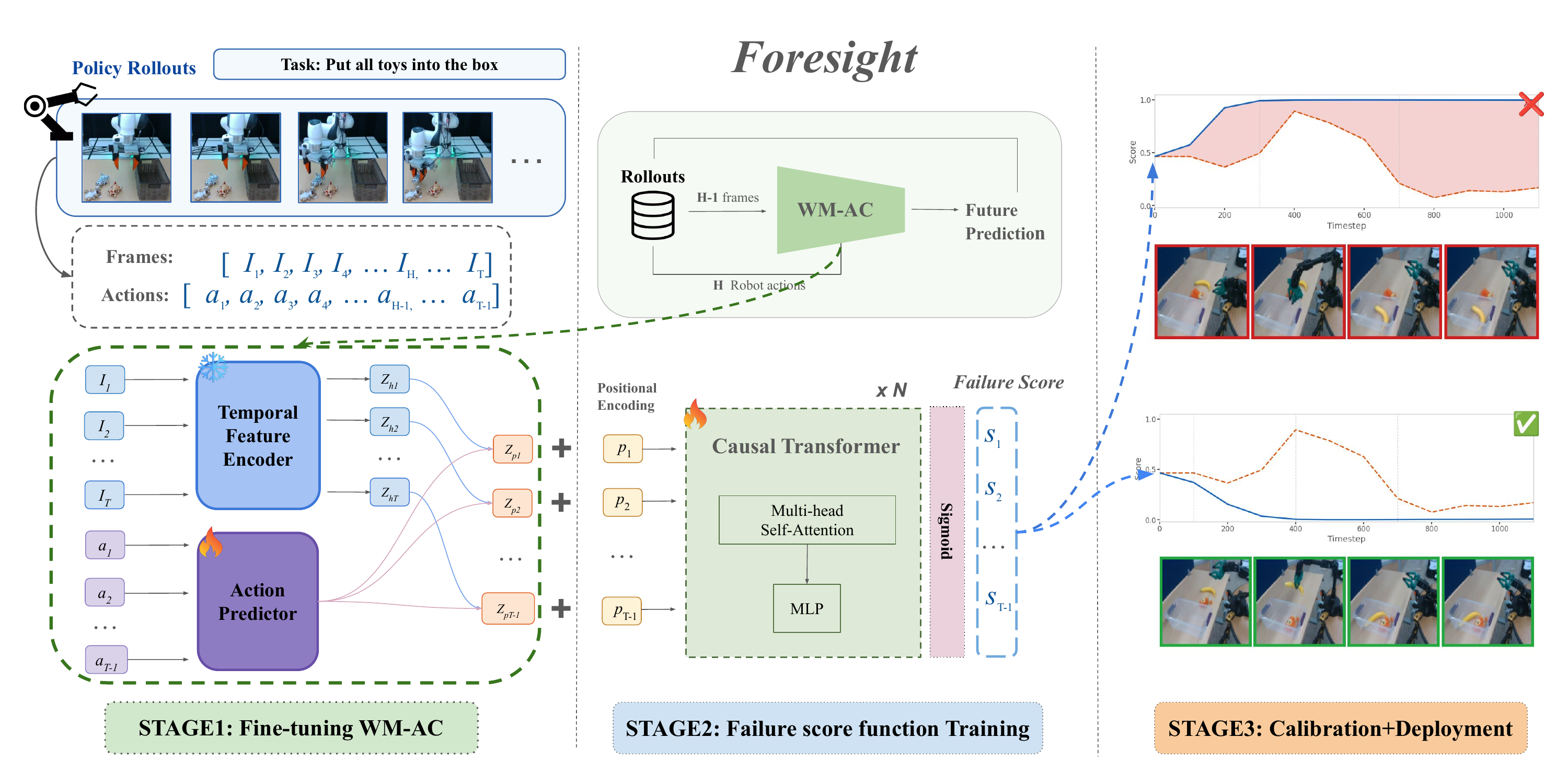}
\vspace{-0.5cm}
\caption{
\textbf{Overview of \ours{}.}
\ours{} consists of three stages.
\textbf{Stage 1:} we fine-tune an action-conditioned world model (WM-AC) on robot rollouts consisting of image observations $I_{1:T}$ and actions $a_{1:T-1}$.
\textbf{Stage 2:} for each timestep $t$, the world model encodes the current observation context into hidden latents $z_t^h$ and predicts action-conditioned future latents $z_t^p$ using the policy-predicted action chunk $A_t$. These latent tokens, together with positional encodings, are passed into a causal sequence model to produce per-timestep failure scores $s_t$.
\textbf{Stage 3:} a conformal calibration set is used to construct a time-varying threshold $\delta_t$ (orange line), and a rollout is flagged as failure once the failure score (blue line) is higher than the threshold $s_t \geq \delta_t$.
}
\vspace{-0.3cm}
\label{fig:pipeline}
\end{figure}

\subsection{System Overview}

We propose \ours{}, a failure detection framework for policies executing long-horizon robotic tasks. Given the current observation context $c_t$ and the policy-predicted action chunk $A_t$, \ours{} uses an action-conditioned world model to extract execution-aware latent features and predicts a per-timestep failure score $s_t$. At deployment, a calibrated time-varying threshold $\delta_t$ converts this score into a binary failure alarm.
Figure~\ref{fig:pipeline} shows the overall pipeline.

Compared with prior failure detection methods that rely on policy-internal features~\citep{gu2025safe,yeh2024aed}, \ours{} uses features from an action-conditioned video world model. This design encourages the detector to capture execution-level failure cues rather than policy-specific artifacts, enabling cross-policy generalization.

\subsection{World Model Feature Extraction}

We use V-JEPA 2-AC~\citep{assran2025vjepa2} as our action-conditioned world model backbone. At each timestep $t$, the world model receives the observation context $c_t$ and the policy-predicted action chunk $A_t$. The temporal feature encoder produces an observed latent representation
\begin{equation}
z_t^h = \mathrm{Pool}\left(f_\phi(c_t)\right),
\end{equation}
where $f_\phi$ is the visual encoder and $\mathrm{Pool}(\cdot)$ averages over spatial patch embeddings. The action predictor then produces an action-conditioned predicted latent
\begin{equation}
z_t^p = \mathrm{Pool}\left(g_\psi(z_t^h, A_t)\right),
\end{equation}
where $g_\psi$ predicts future latent states conditioned on the proposed action chunk.

The hidden latent $z_t^h$ captures what is currently observed, while the predicted latent $z_t^p$ captures what the world model expects to happen under the policy's next action chunk. We use the predicted latent to form the timestep token

\begin{equation}
u_t = W z_t^p + p_t,
\end{equation}
where $p_t \in \mathbb{R}^d$ is a fixed sinusoidal positional encoding. We compare $z_t^p$ and  $z_t^h$ as inputs in Appendix~\ref{app:hidden_vs_pred} to validate the importance of action conditioning in feature selection.

\subsection{Failure Scoring with Causal Sequence Models}

Given latent tokens up to timestep $t$,
\begin{equation}
U_{\leq t} = \{u_1, u_2, \dots, u_t\},
\end{equation}
we use a causal sequence model to predict a per-timestep failure score:
\begin{equation}
s_t = D_\theta(U_{\leq t}) \in [0,1].
\end{equation}
The causal mask ensures that the detector only uses information available up to the current timestep. We implement $D_\theta$ using a causal Transformer with positional encodings and masked self-attention, and compare it against MLP and LSTM variants in the experiments.

The detector is trained using trajectory-level binary labels. Since failure timestamps are not annotated, each timestep inherits the rollout-level label, with early-detection weighting applied to encourage high scores before or during failure events.

\subsection{Conformal Prediction Thresholding}

Following previous works~\citep{xu2025can, ho2026worldmodel, gu2025safe},
we adopt functional conformal prediction
(FCP)~\citep{diquigiovanni2025importance} to convert the continuous
failure score $s_t$ into a binary alarm with statistical guarantees.
FCP constructs a one-sided time-varying upper band
$\delta_t = \mu_t + h_t$ calibrated on successful rollouts from a
calibration split, where $\mu_t$ is the time-varying mean score and
$h_t$ is a calibrated bandwidth term. We provide the detailed construction
of $h_t$ in Appendix~\ref{app:cp-alpha}.

A failure is declared at the first step where the score exceeds the band:
\begin{equation}
    \hat{y}_t = \mathbf{1}[s_t \geq \delta_t].
\end{equation}
Under mild exchangeability assumptions~\citep{vovk2005algorithmic}, this
guarantees that the false positive rate, i.e., the probability of flagging
a truly successful rollout as a failure at any point during execution, is
controlled at level $\alpha$. We evaluate across a range of significance
levels $\alpha$ in Section~\ref{sec:exp}.

\section{Experiment}
\label{sec:exp}

\subsection{Evaluation Benchmarks}
\label{sec:eval_benchmarks}

We evaluate our method on three long-horizon manipulation benchmark suites:
LIBERO-Long~\citep{liu2023libero}, ManiSkill-Long~\citep{taomaniskill3}, and BEHAVIOR-1K~\citep{li2024behavior1k}. These benchmarks cover different task horizons, environment complexity, robot embodiments, and policy sources. Table~\ref{tab:benchmark_summary} summarizes the main properties of each benchmark.
\vspace{-4mm}
\paragraph{LIBERO-Long}
LIBERO-Long is a widely used benchmark for evaluating vision-language-action policies on long-horizon tabletop manipulation. Its tasks typically require multiple object-interaction steps, such as placing two objects into a target container. We evaluate OpenVLA~\citep{kim24openvla} and $\pi_0$-FAST~\citep{black2024pi0} on LIBERO-Long using their officially released checkpoints.
\vspace{-4mm}

\paragraph{ManiSkill-Long}
To evaluate longer and more compositional manipulation behaviors, we construct four tasks in ManiSkill~\citep{taomaniskill3}, referred to as ManiSkill-Long. These tasks require at least eight symbolic subgoals to accomplish. For example, \textit{stack\_6\_cube} requires the robot to sequentially stack six cubes, which involves 12 pick-and-place actions. We evaluate $\pi_0$-FAST~\citep{black2024pi0} using self-collected rollouts, where the policy is fine-tuned from the corresponding checkpoints using our collected data.
\vspace{-4mm}

\paragraph{BEHAVIOR-1K}
We further evaluate on four mobile manipulation tasks selected from the BEHAVIOR-1K challenge long-horizon benchmark. Unlike LIBERO-Long and ManiSkill-Long, BEHAVIOR-1K requires both navigation and manipulation in larger household environments. We evaluate a revised version of $\pi_{0.5}$~\citep{black2025pi05} based on the best solution~\citep{larchenko2025behavior} from the BEHAVIOR-1K challenge.
\vspace{-4mm}

\paragraph{Real-World Experiment}
Beyond simulation, we validate our approach in real-world manipulation experiments. 
We collect rollouts via teleoperation and evaluate three policies, ACT~\citep{zhao2023learning}, $\pi_{0.5}$~\citep{black2025pi05}, and SmolVLA~\citep{shukor2025smolvla}, on a ReactorX-200 arm across three tabletop arrangement tasks. 
To assess cross-embodiment generalization, we additionally evaluate GR00T N1.5~\citep{gr00tn1_2025} on a toy pick-up task using a Franka arm.

Additional task-level details, rollout statistics, and policy information are provided in Appendix~\ref{app:benchmark_details}.

\begin{table*}[t]
\centering
\small
\vspace{-0.2cm}
\setlength{\tabcolsep}{4pt}
\begin{tabularx}{\textwidth}{lccXc}
\toprule
Benchmark 
& \#Tasks 
& Embodiment 
& Evaluated Policies 
& Avg. Steps  \\
\midrule
LIBERO-Long~\citep{liu2023libero}
& 10 
& Franka 
& OpenVLA/$\pi_0$-FAST 
& 253 \\

ManiSkill-Long~\citep{taomaniskill3}
& 4 
& Franka 
& $\pi_0$-FAST 
& 1,484 \\

BEHAVIOR-1K~\citep{li2024behavior1k} 
& 4 
& R1Pro 
& revised $\pi_{0.5}$ 
& 8,557 \\

Real-World Exp
& 4 
& ReactorX/Franka 
& ACT/$\pi_{0.5}$/GR00T N1.5/SmolVLA
& 1,175 \\

\bottomrule
\end{tabularx}
\vspace{-0.3cm}
\caption{
Summary of the evaluation benchmarks. Each row corresponds to one benchmark suite. Average simulation steps are computed over successful rollouts; when multiple evaluated policies have different rollout horizons, we report the observed range.
}
\vspace{-0.2cm}
\label{tab:benchmark_summary}
\end{table*}

\subsection{Baselines}

\label{sec:baselines}

We compare \ours against four representative runtime failure detection baselines. Rollout-level ROC-AUC and balanced accuracy are computed using the protocol in Section~\ref{sec:eval_metrics}. 
When a baseline requires a detection threshold, we calibrate it using the same held-out successful rollouts and sweep over the same significance levels $\alpha$. We report the best-performing variant within each baseline:

FAIL-Detect~\citep{xu2025can} is an uncertainty-based OOD detection method using only successful rollouts.
\textit{SAFE}~\citep{gu2025safe} trains on both success and failure rollouts. Predictions are calibrated with the same functional conformal prediction procedure used for \ours. \textit{SAFE} serves as the policy-internal-representation baseline.
RND~\citep{he2024rediffuser} models the embedding distribution of successful rollouts for OOD detection.
Gauge~\citep{ho2026worldmodel} uses compressed video world-model latents together with conformal decision functions to classify executions as success, known failure, or out-of-distribution anomaly.
We adapt it to our binary failure detection setting by collapsing all non-success outputs into failures and training on success data only. This baseline compares \ours against a recent world-model approach that uses video latents and conformal thresholding.

\subsection{Evaluation Metrics}
\label{sec:eval_metrics}

Following the evaluation protocol of previous work~\citep{gu2025safe}, we assess rollout-level failure prediction using ROC-AUC and balanced accuracy, which respectively measure \textbf{threshold-independent} score separability and \textbf{threshold-dependent} classification performance.

\paragraph{ROC-AUC}
Given a per-timestep failure score $s_t$, we aggregate it into a rollout-level score by taking the maximum value over the trajectory:
\begin{equation}
    \bar{s} = \max_{t=1,\ldots,T} s_t.
\end{equation}
We then compute ROC-AUC using $\bar{s}$ to evaluate how well the score separates failed rollouts from successful ones across all possible thresholds. A higher ROC-AUC indicates stronger threshold-independent discriminative ability.

\paragraph{Balanced accuracy}
For threshold-based evaluation, each rollout is classified as successful or failed according to the selected detection threshold. We report balanced accuracy,
\begin{equation}
    \mathrm{BalAcc} = \frac{1}{2}(\mathrm{TPR} + \mathrm{TNR}),
\end{equation}
where TPR denotes the true positive rate and TNR denotes the true negative rate. Balanced accuracy assigns equal weight to successful and failed rollouts, making it robust to class imbalance.

We evaluate all baselines and our method using 3-fold cross-validation. In experiments, we sweep $\alpha$ to evaluate multiple operating points and report the value that gives the best cross-validation balanced accuracy. To ensure conclusions are not driven only by this threshold choice, we also report ROC-AUC, which evaluates threshold-independent score separability.

\subsection{Experiment Results}

\label{sec:experiment_results}

\paragraph{Simulation failure detection.}
Table~\ref{tab:main_results} compares \ours{} with baselines on the three simulated benchmarks. Across all three datasets, the strongest \ours{}-Transformer achieves the best calibrated balanced accuracy: $0.94 \pm 0.06$ on LIBERO-Long, $0.80 \pm 0.10$ on ManiSkill-Long, and $0.78 \pm 0.02$ on BEHAVIOR-1K. On the two longer-horizon benchmarks, \ours-Transformer also obtains the best threshold-independent ROC-AUC, reaching $0.84 \pm 0.03$ on ManiSkill-Long and $0.76 \pm 0.02$ on BEHAVIOR-1K. 

The gains are most pronounced on BEHAVIOR-1K, the longest and most challenging benchmark in our evaluation. As summarized in Table~\ref{tab:benchmark_summary}, BEHAVIOR-1K rollouts average $8{,}557$ simulation steps, roughly $34\times$ longer than LIBERO-Long and $5.8\times$ longer than ManiSkill-Long. This setting requires detecting failures over extended executions rather than short rollouts, and it goes beyond the horizons for which the baseline methods were originally designed and evaluated. In this regime, the best non-\ours{} baseline reaches $0.72 \pm 0.02$ ROC-AUC and $0.64 \pm 0.05$ balanced accuracy, whereas \ours-Transformer reaches $0.76 \pm 0.02$ ROC-AUC and $0.78 \pm 0.02$ balanced accuracy. This $0.14$ balanced-accuracy and $0.04$ ROC-AUC improvement suggests that action-conditioned world-model features are especially useful over a long trajectory.

\begin{figure}[t]
    \centering
    \vspace{-0.4cm}
    \includegraphics[width=0.95\linewidth]{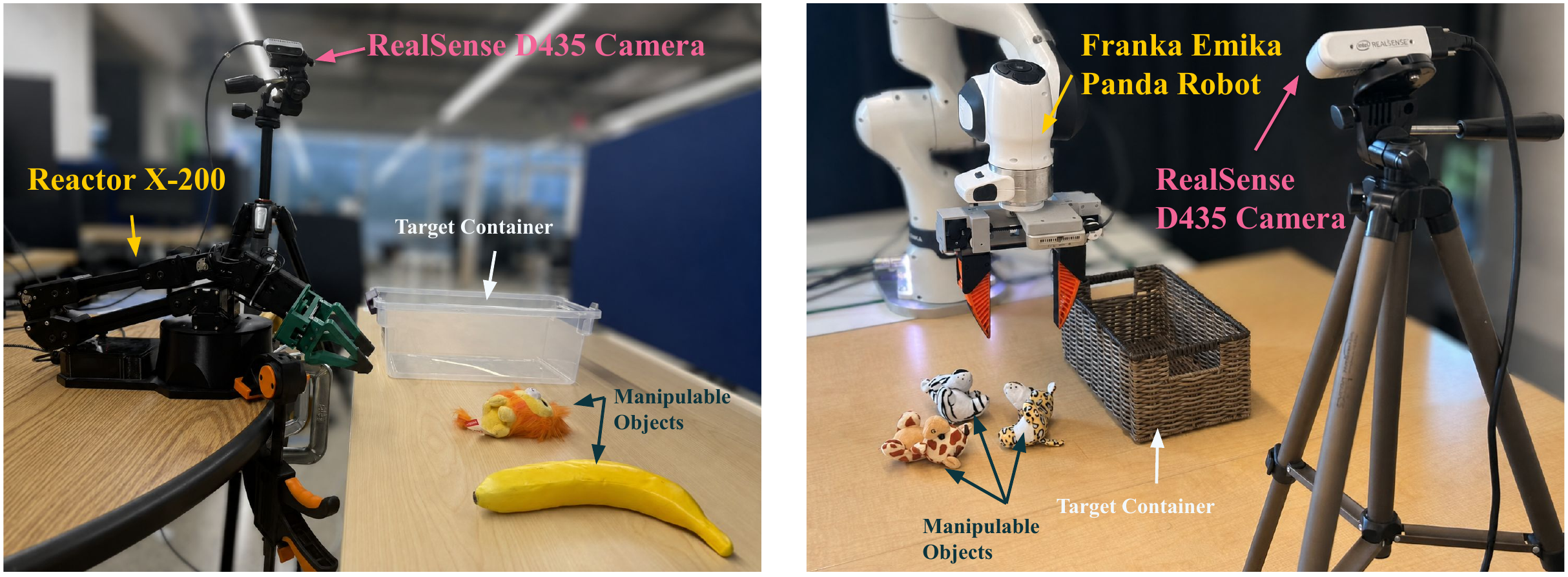}
    \vspace{-0.2cm}
    \caption{Real-Robot Setup. \textbf{Left:} real-world robot setting for three table-top manipulation tasks using ReactorX-200 arm. 
    \textbf{Right:} real-world robot setting for a three-toy picking task using Franka arm.
    }
    \label{fig:real_setup}
\end{figure}

\definecolor{bestnavy}{HTML}{1F4E79}
\definecolor{secondteal}{HTML}{4F8A8B}
\begin{table*}[t]
\centering
\small
\setlength{\tabcolsep}{4pt}
\begin{tabular}{lcccccc}
\toprule
Method
& \multicolumn{2}{c}{LIBERO-Long}
& \multicolumn{2}{c}{ManiSkill-Long}
& \multicolumn{2}{c}{BEHAVIOR-1K} \\
\cmidrule(lr){2-3}\cmidrule(lr){4-5}\cmidrule(lr){6-7}
& ROC-AUC & BalAcc & ROC-AUC & BalAcc & ROC-AUC & BalAcc \\
\midrule
FAIL-Detect~\citep{xu2025can} 
& \textcolor[HTML]{D55E00}{0.90 $\pm$ 0.02} & 0.82 $\pm$ 0.06 
& 0.71 $\pm$ 0.02 & 0.50 $\pm$ 0.01
& 0.54 $\pm$ 0.06 & 0.52 $\pm$ 0.01 \\
SAFE-MLP~\citep{gu2025safe} 
& 0.52 $\pm$ 0.01 & 0.50 $\pm$ 0.01 
& 0.61 $\pm$ 0.02 & 0.53 $\pm$ 0.02
& 0.50 $\pm$ 0.00 & 0.50 $\pm$ 0.00 \\
SAFE-LSTM~\citep{gu2025safe} 
& \textbf{\textcolor[HTML]{0072B2}{0.91 $\pm$ 0.02}} & 0.88 $\pm$ 0.02 
& 0.82 $\pm$ 0.01 & 0.74 $\pm$ 0.01 
& 0.72 $\pm$ 0.02 & 0.64 $\pm$ 0.05 \\
RND~\citep{he2024rediffuser} 
& \textcolor[HTML]{D55E00}{0.90 $\pm$ 0.02} & 0.83 $\pm$ 0.04 
& \textcolor[HTML]{D55E00}{0.83 $\pm$ 0.02} & 0.68 $\pm$ 0.18
& 0.65 $\pm$ 0.01 & 0.54 $\pm$ 0.04\\
Gauge~\citep{ho2026worldmodel} 
& 0.88 $\pm$ 0.01 & 0.81 $\pm$ 0.06 
& 0.80 $\pm$ 0.02 & 0.77 $\pm$ 0.03
& 0.61 $\pm$ 0.03 & 0.60 $\pm$ 0.03 \\
\midrule
\ours-MLP 
& 0.88 $\pm$ 0.01 & 0.80 $\pm$ 0.02 
& 0.70 $\pm$ 0.03 & 0.71 $\pm$ 0.18 
& 0.73 $\pm$ 0.02 & 0.56 $\pm$ 0.03 \\
\ours-LSTM 
& 0.86 $\pm$ 0.02 & \textcolor[HTML]{D55E00}{0.89 $\pm$ 0.03} 
& 0.76 $\pm$ 0.00 & \textcolor[HTML]{D55E00}{0.79 $\pm$ 0.16} 
& \textcolor[HTML]{D55E00}{0.75 $\pm$ 0.04} & \textcolor[HTML]{D55E00}{0.75 $\pm$ 0.09} \\
\ours-Transformer 
& 0.89 $\pm$ 0.02 & \textbf{\textcolor[HTML]{0072B2}{0.94 $\pm$ 0.06} }
& \textbf{\textcolor[HTML]{0072B2}{0.84 $\pm$ 0.03}} & \textbf{\textcolor[HTML]{0072B2}{0.80 $\pm$ 0.10}} 
& \textbf{\textcolor[HTML]{0072B2}{0.76 $\pm$ 0.02}} & \textbf{\textcolor[HTML]{0072B2}{0.78 $\pm$ 0.02}} \\
\bottomrule
\end{tabular}
\vspace{-0.2cm}
\caption{
Main rollout-level failure detection results. Values are reported as mean $\pm$ standard deviation across folds and rounded to two decimals. ROC-AUC is computed from the maximum failure score over each rollout on the test split. Balanced accuracy is computed after calibrating the detection threshold on the calibration split and selecting the tuned $\alpha$. Best results are shown in bold blue, and second-best results are shown in orange. All \ours{} methods use action-conditioned latent predictions. LIBERO-Long and ManiSkill-Long use $\pi_0$-FAST rollouts~\citep{black2024pi0,pertsch2025fast}, while BEHAVIOR-1K uses rollouts from a $\pi_{0.5}$ model revised for the best BEHAVIOR-1K solution~\citep{larchenko2025behavior}. Gauge reports the best performance among its seven methods. Full experiment results with the selected alpha values are provided in Appendix~\ref{app:cp-alpha}.
}

\label{tab:main_results}
\end{table*}

\newcommand{\best}[1]{\textbf{\textcolor[HTML]{0072B2}{#1}}}
\newcommand{\second}[1]{\textcolor[HTML]{D55E00}{#1}}

\begin{table*}[t]
\centering
\small
\setlength{\tabcolsep}{5pt}
\vspace{-0.3cm}
\begin{tabular}{lcccc}
\toprule
Method & ReactorX / ACT & ReactorX / $\pi_{0.5}$ & ReactorX / SmolVLA & Franka / GR00T N1.5 \\
\midrule
FAIL-Detect~\citep{xu2025can}
  & $0.85 \pm 0.07$
  & $0.64 \pm 0.06$
  & $0.71 \pm 0.05$
  & \second{$0.88 \pm 0.05$} \\
SAFE-MLP~\citep{gu2025safe}
  & \second{$0.89 \pm 0.05$}
  & $0.66 \pm 0.36$
  & $0.64 \pm 0.19$
  & $0.50 \pm 0.10$ \\
SAFE-LSTM~\citep{gu2025safe}
  & $0.70 \pm 0.07$
  & $0.75 \pm 0.14$
  & $0.43 \pm 0.10$
  & $0.79 \pm 0.10$ \\
RND~\citep{he2024rediffuser}
  & $0.86 \pm 0.04$
  & $0.78 \pm 0.06$
  & \best{$0.82 \pm 0.03$}
  & $0.64 \pm 0.15$ \\
\midrule
\ours-MLP
  & $0.50 \pm 0.00$
  & $0.55 \pm 0.05$
  & $0.53 \pm 0.22$
  & $0.59 \pm 0.20$ \\
\ours-LSTM
  & $0.85 \pm 0.05$
  & \second{$0.85 \pm 0.03$}
  & $0.64 \pm 0.08$
  & $0.66 \pm 0.08$ \\
\ours-Transformer
  & \best{$0.93 \pm 0.01$}
  & \best{$0.87 \pm 0.03$}
  & \second{$0.79 \pm 0.09$}
  & \best{$0.89 \pm 0.10$} \\
\bottomrule
\end{tabular}
\vspace{-0.35cm}
\caption{
  Real-world manipulation results. We collect teleoperated rollouts and
  evaluate rollout-level failure detection across
  ACT~\citep{zhao2023learning}, $\pi_{0.5}$~\citep{black2025pi05}, and
  SmolVLA~\citep{shukor2025smolvla} on a ReactorX arm over tabletop
  arrangement tasks. To assess cross-embodiment generalization, we
  further evaluate GR00T~N1.5~\citep{gr00tn1_2025} on a toy pick-up
  task using a Franka arm. We compare \ours{} against the same baselines
  as in the simulation experiments and report ROC-AUC as mean $\pm$
  standard deviation. Blue indicates the best result and
  orange indicates the second-best result in each column.
}
\label{tab:real_world_results}
\end{table*}

\begin{table*}[!htpb]
\centering
\small
\setlength{\tabcolsep}{5pt}
\begin{tabularx}{\textwidth}{l X X c c}
\toprule
Benchmark & Train distribution & Test distribution & ROC-AUC & BalAcc \\
\midrule
LIBERO-Long 
  & $\pi_0$-FAST rollouts 
  & OpenVLA rollouts 
  & $0.64\pm0.02$ 
  & $0.90\pm0.01$ \\
Real-World Exp.
  & $\pi_{0.5}$ rollouts
  & ACT rollouts
  & $0.94\pm0.02$ 
  & $0.82\pm0.08$ \\
Real-World Exp.
  & ACT rollouts
  & $\pi_{0.5}$ rollouts
  & $0.56\pm0.07$ 
  & $0.52\pm0.03$ \\
Real-World Exp.
  & SmolVLA rollouts 
  & ACT rollouts
  & $0.92\pm0.04$  
  & $0.73\pm0.07$  \\
Real-World Exp.
  & $\pi_{0.5}$ rollouts 
  & SmolVLA rollouts
  & $0.67\pm0.02$ 
  & $0.62\pm0.01$  \\
\bottomrule
\end{tabularx}
\vspace{-0.3cm}
\caption{
Generalization experiments. Cross-policy transfer evaluates whether the detector learns execution-level failure cues rather than policy-specific artifacts.
}
\vspace{-0.3cm}
\label{tab:generalization_results}
\end{table*}

\paragraph{Real-world rollout monitoring.}
Table~\ref{tab:real_world_results} shows that \ours transfers to real-robot rollout monitoring across policies and embodiments. \ours-Transformer achieves the best ROC-AUC in three of the four settings: ReactorX / ACT ($0.93 \pm 0.01$), ReactorX / $\pi_{0.5}$ ($0.87 \pm 0.03$), and Franka / GR00T N1.5 ($0.89 \pm 0.10$). Across settings, \ours-LSTM is also consistently strong, whereas \ours-MLP remains near chance ($0.50$--$0.59$). These results suggest that action-conditioned world model features are useful for real-world failure detection, but robust rollout monitoring requires sequence-level detectors rather than independent frame-level classification.

\paragraph{Cross-policy generalization.}
Table~\ref{tab:generalization_results} evaluates whether a detector trained on one policy distribution can transfer to another. The results show that cross-policy generalization is feasible with \ours{}, as detectors trained on $\pi_0$-FAST or $\pi_{0.5}$ can successfully transfer to different test policies. However, transfer is policy-dependent and can be asymmetric. In the real-world setting, training on $\pi_{0.5}$ transfers well to ACT, while training on ACT transfers poorly to $\pi_{0.5}$. One possible reason is that $\pi_{0.5}$ rollouts contain broader behaviors, including recovery trajectories. For example, in a sequential task where the policy should pick the lion first and the banana second, ACT or SmolVLA may fail after missing the lion, whereas $\pi_{0.5}$ may recover by picking the banana and then returning to pick the lion. A detector trained only on ACT-like rollouts may not see such recovery behavior and may misclassify it as failure. Overall, these results suggest that \ours{} can generalize across policies, but the strength of transfer depends on whether the training policy covers the behaviors and failure modes of the target policy.

\section{Conclusion}
\label{sec:conclusion}

We presented \ours, a failure detection framework for long-horizon robotic manipulation that monitors rollouts using action-conditioned world-model representations. By combining V-JEPA-style latent prediction with causal failure detectors and functional conformal calibration, \ours detects failures using only trajectory-level success/failure labels and does not require access to policy-internal states or uncertainty estimates. Across LIBERO-Long, ManiSkill-Long, BEHAVIOR-1K, and real-robot experiments, our results show that action-conditioned predicted latents provide effective signals for identifying execution failures, particularly on longer-horizon tasks. These findings suggest that action-grounded world-model embeddings are a promising interface for scalable and policy-adaptable runtime monitoring in robotic manipulation.

\paragraph{Limitations}
A key limitation is the computational cost and latency of pretrained world models, which makes on-device deployment challenging and may limit applicability to highly reactive or agile tasks requiring fast closed-loop control. In addition, while conformal calibration helps control false alarms under held-out successful rollouts, its guarantees depend on the calibration distribution matching deployment conditions.
\bibliography{reference}  
\clearpage

\section*{Appendix}
\label{sec:app}

\section{More Implementation Details}
\label{app:implementation}

\paragraph{World-model feature extraction.}
We use V-JEPA~2-AC~\citep{assran2025vjepa2} as the action-conditioned world-model backbone,
initialized from the pretrained \texttt{vjepa2-ac-vitg.pt} checkpoint (ViT-Giant encoder).
The visual encoder is frozen throughout; only the action-conditioned predictor is trained
on robot rollouts from the corresponding benchmark.
Images are resized to $256\!\times\!256$ and normalized with ImageNet statistics.
The encoder uses a patch size of $16\!\times\!16$ with tubelet size~2,
yielding 256 spatial patch tokens per frame.
At each replanning step, the model receives a sliding window of 8 frames
(non-overlapping) as the observation context, together with the policy-predicted action chunk, whose action dimensionality depends on the benchmark and robot embodiment: 7D for LIBERO and the real-world ACT/$\pi_{0.5}$ policies, 8D for ManiSkill-Long, 10D for the real-world Franka setup, and 23D for the BEHAVIOR-1K R1Pro robot.
The predictor has 24 transformer layers with embedding dimension 1024
and 16 attention heads, and is frame-causal (\texttt{pred\_is\_frame\_causal=True}).
Its output patch tokens are mean-pooled over all 256 spatial patches to
produce a 1408-dimensional latent vector per frame (matching the ViT-Giant
encoder embedding dimension), which is passed to the failure detector.

\paragraph{Action-conditioned predictor training.}
The predictor is trained with a combined teacher-forcing and
autoregressive-rollout objective following the V-JEPA~2-AC training
procedure~\citep{assran2025vjepa2}.
Briefly, at each iteration the predictor is run in teacher-forcing mode
(ground-truth target-encoder features as context) and in autoregressive
rollout mode ($n{=}2$ steps), and the L1 losses on LayerNorm-normalized
representations are summed:
$\mathcal{L} = \mathcal{L}_{\mathrm{TF}} + \mathcal{L}_{\mathrm{AR}}$.
We use AdamW ($\beta_1{=}0.9$, $\beta_2{=}0.999$) with weight decay
0.04, a linear LR warmup over 10 epochs, and cosine annealing to 0
for 200 epochs total.
The visual encoder remains frozen, and the predictor is trained from scratch for all experiments.
For LIBERO we train on a single H200 GPU with batch size 256 and peak
LR $2{\times}10^{-4}$ (warmup from $2.5{\times}10^{-5}$).
For BEHAVIOR-1K and Maniskill-Long we use $2{\times}$H200 GPUs with
effective batch size 512 and the same LR schedule.
For real-world benchmarks (Franka, ACT, $\pi_{0.5}$, SmolVLA) we use
$2{\times}$H200 GPUs with effective batch size 32 (16 per GPU) and
peak LR $5{\times}10^{-5}$ (warmup from $5{\times}10^{-6}$).

\paragraph{Failure detector architectures.}
We evaluate three detector architectures on V-JEPA~2-AC features:
\textbf{MLP}, \textbf{LSTM}, and \textbf{causal Transformer}.
All three share the following hyperparameters:
input dimension 1408 (the world-model latent),
2 layers, hidden dimension 256, learning rate $10^{-4}$ (Adam),
$\ell_2$ regularization $\lambda{=}10^{-2}$, dropout 0.1, and 300 training epochs
with batch size 512 on a single H200 GPU.

The \textbf{MLP} projects the input through two linear layers
(Linear$\to$ReLU$\to$Linear$\to$Sigmoid), treating each timestep independently.

The \textbf{LSTM} is a 2-layer LSTM with hidden dimension 256,
followed by a Linear$\to$Sigmoid output head.
It processes the full episode sequence with dropout applied
between layers and on the final hidden state.

The \textbf{causal Transformer} applies a learned linear projection
to dimension 256, adds sinusoidal positional encodings,
then passes through 2 pre-norm TransformerEncoder layers with 4 attention heads,
feedforward dimension 1024 ($=4\!\times\!256$), and dropout 0.1.
A causal attention mask ensures that the score at timestep~$t$
depends only on features up to and including~$t$.
A final Linear$\to$Sigmoid head produces per-step failure probabilities.

\section{Data Splits and Calibration Protocol}
\label{app:calibration_protocol}

We first randomly shuffle all rollouts and partition them into three equal-sized folds, yielding three experimental rounds in accordance with the standard 3-fold cross-validation protocol. In each round, one fold is held out as the test set, while the remaining two folds are used for model development. Specifically, these two folds are further split into a training set, validation set, and calibration set in a 6:1:1 ratio. The training set is used to fit the downstream detector, the validation set is used for model selection and hyperparameter tuning, and the calibration set is used to construct time-varying conformal thresholds. For the AC predictor, however, we use all non-test data available in each round, including the training, validation, and calibration sets. We assume the AC predictor has full access to all data except the held-out test set.

\section{Conformal Prediction Thresholding}
\definecolor{bestnavy}{HTML}{1F4E79}
\definecolor{secondteal}{HTML}{4F8A8B}

\begin{table*}[t]
\centering
\small
\setlength{\tabcolsep}{4pt}
\begin{tabular}{lcccccc}
\toprule
Method
& \multicolumn{2}{c}{LIBERO-Long}
& \multicolumn{2}{c}{ManiSkill-Long}
& \multicolumn{2}{c}{BEHAVIOR-1K} \\
\cmidrule(lr){2-3}\cmidrule(lr){4-5}\cmidrule(lr){6-7}
& BalAcc & Best $\alpha$
& BalAcc & Best $\alpha$
& BalAcc & Best $\alpha$ \\
\midrule
FAIL-Detect~\citep{xu2025can}
& $0.82 \pm 0.06$ & $0.10$
& $0.50 \pm 0.01$ & $0.15$
& $0.52 \pm 0.01$ & $0.02$ \\
SAFE-MLP~\citep{gu2025safe}
& $0.50 \pm 0.01$ & $0.02$
& $0.53 \pm 0.02$ & $0.20$
& $0.50 \pm 0.00$ & $0.02$ \\
SAFE-LSTM~\citep{gu2025safe}
& $0.88 \pm 0.02$ & $0.02$
& $0.74 \pm 0.01$ & $0.25$
& $0.64 \pm 0.05$ & $0.10$ \\
RND~\citep{he2024rediffuser}
& $0.83 \pm 0.04$ & $0.02$
& $0.68 \pm 0.18$ & $0.05$
& $0.54 \pm 0.04$ & $0.02$ \\
Gauge~\citep{ho2026worldmodel}
& $0.81 \pm 0.06$ & $0.20$
& $0.77 \pm 0.03$ & $0.25$
& $0.60 \pm 0.03$ & $0.20$ \\
\midrule
\ours{}-MLP
& $0.80 \pm 0.02$ & $0.10$
& $0.71 \pm 0.18$ & $0.15$
& $0.56 \pm 0.03$ & $0.02$ \\
\ours{}-LSTM
& $0.89 \pm 0.03$ & $0.05$
& $0.79 \pm 0.16$ & $0.02$
& $0.75 \pm 0.09$ & $0.02$ \\
\ours{}-Transformer
& $0.94 \pm 0.06$ & $0.02$
& $0.80 \pm 0.10$ & $0.02$
& $0.78 \pm 0.02$ & $0.20$ \\
\bottomrule
\end{tabular}
\vspace{-0.2cm}
\caption{
Balanced accuracy and selected $\alpha$ on simulation benchmarks.
For each method and benchmark, $\alpha$ is chosen by maximizing
balanced accuracy over the candidate set using 3-fold cross-validation.
}
\label{tab:best_alpha_sim}
\end{table*}
\label{app:cp-alpha}

As described in the main text, FCP constructs a one-sided time-varying
upper threshold
\begin{equation}
    \delta_t = \mu_t + h_t,
\end{equation}
where $\mu_t$ is the mean score trajectory estimated from successful
calibration rollouts and $h_t$ is a calibrated bandwidth term. We now
describe how $h_t$ is instantiated.

Let $\{s^{(i)}_t\}_{i=1}^{n}$ denote the score trajectories of the
$n$ successful rollouts in a held-out calibration set. We estimate the
mean trajectory as
\begin{equation}
    \mu_t = \frac{1}{n}\sum_{i=1}^{n} s^{(i)}_t.
\end{equation}
We further estimate a time-varying modulation term $\sigma_t$, which
captures how scores deviate from the mean across calibration trajectories.
For each calibration rollout, we compute the normalized nonconformity score
\begin{equation}
    R_i = \sup_t \frac{s^{(i)}_t - \mu_t}{\sigma_t}.
\end{equation}
Let $\hat{q}$ be the $(1-\alpha)$-quantile of the calibration
nonconformity scores $\{R_i\}_{i=1}^{n}$. The bandwidth term in the main
text is then given by
\begin{equation}
    h_t = \hat{q}\sigma_t,
\end{equation}
which yields the time-varying threshold
\begin{equation}
    \delta_t = \mu_t + h_t = \mu_t + \hat{q}\sigma_t.
\end{equation}

A failure alarm is declared at the first step where the score exceeds the
band:
\begin{equation}
    \hat{y}_t = \mathbf{1}[s_t \geq \delta_t].
\end{equation}

\paragraph{Selection of $\alpha$.}
We sweep a fixed candidate set
\[
\alpha \in \{0.02,\,0.05,\,0.10,\,0.15,\,0.20,\,0.25,\,0.30,\,0.35,
             0.40,\,0.45,\,0.50,\,0.60,\,0.70,\,0.80,\,0.90\}.
\]
For each value of $\alpha$, the time-varying threshold $\delta_t$ is computed
solely from the dedicated calibration split.
The operating $\alpha$ is selected per method and benchmark by maximizing
balanced accuracy aggregated across three cross-validation folds;
the selected value is then fixed before reporting test results.
Table~\ref{tab:best_alpha_sim} reports
the selected $\alpha$ alongside balanced accuracy for simulation.

\section{Baseline Implementation Details}
\label{app:baseline_details}
\vspace{-0.4cm}
Baseline details are in Table~\ref{tab:baseline_summary}. To ensure fair comparison, all baselines are evaluated using the same train/calibration/test splits whenever applicable. 
When a method requires thresholding, we calibrate it using the same held-out rollouts used for \ours{}.

For Gauge, we use the authors' released code and default hyperparameters, but we use only the success model because our setting contains success and failure labels but no separate OOD split, and the original paper reports that the success model outperforms the failure model. Thus, for this baseline, we train and calibrate only on successful rollouts. Gauge also reports multiple CP scoring methods, such as reconstruction error and latent distance (L2), so we report the best-performing variant for each dataset.

\begin{table*}[!htbp]
\centering
\small
\begin{tabular}{lcccc}
\toprule
Method & Input signal & Uses failures for training? & Uses policy internals?\\
\midrule
FAIL-Detect$^\dagger$ & VLA internal latent  & No & Yes  \\
RND$^\dagger$         & VLA internal latent + action predictions & No & Yes  \\
SAFE-MLP & VLA internal latent & Yes & Yes \\
SAFE-LSTM & VLA internal latent & Yes & Yes  \\
Gauge & World-model video latents & No & No \\
\ours{} & Action-conditioned world-model latents & Yes & No \\
\bottomrule
\end{tabular}
\caption{
Baseline comparison summary. 
This table clarifies which information each method is allowed to use and how thresholds are calibrated.
$^\dagger$We adapt the original method by replacing its image observation input with the VLA's internal latent, making it directly comparable with the other policy-internal baselines.
}
\label{tab:baseline_summary}
\end{table*}

\clearpage

\section{Additional Benchmark Details}
\label{app:benchmark_details}

\begin{figure}[!htbp]
    \centering
    \includegraphics[width=1\linewidth]{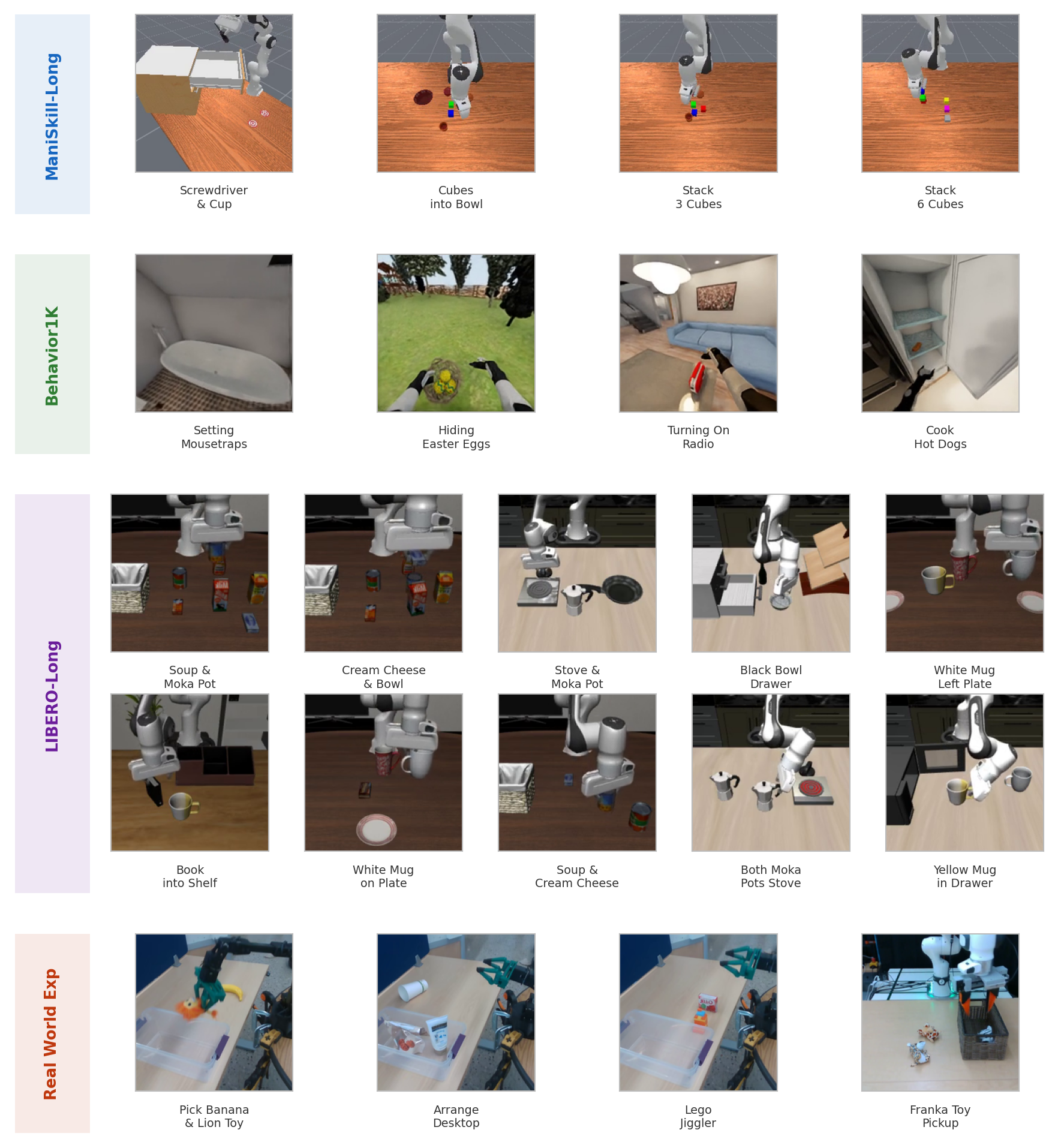}
    \caption{Benchmark tasks overview}
    \label{fig:benchmark_task_overview}
\end{figure}

This section provides additional details for the benchmark suites used in our experiments, including task names, number of rollouts, success rates of the evaluated policies, and rollout horizon statistics. 
These details complement the benchmark summary in Section~\ref{sec:eval_benchmarks}.
A visualization can be found in Fig.~\ref{fig:benchmark_task_overview}.
\clearpage

\subsection{LIBERO-Long}
\label{app:libero_details}

\begin{figure}[b!]
    \centering
    \includegraphics[width=1\linewidth]{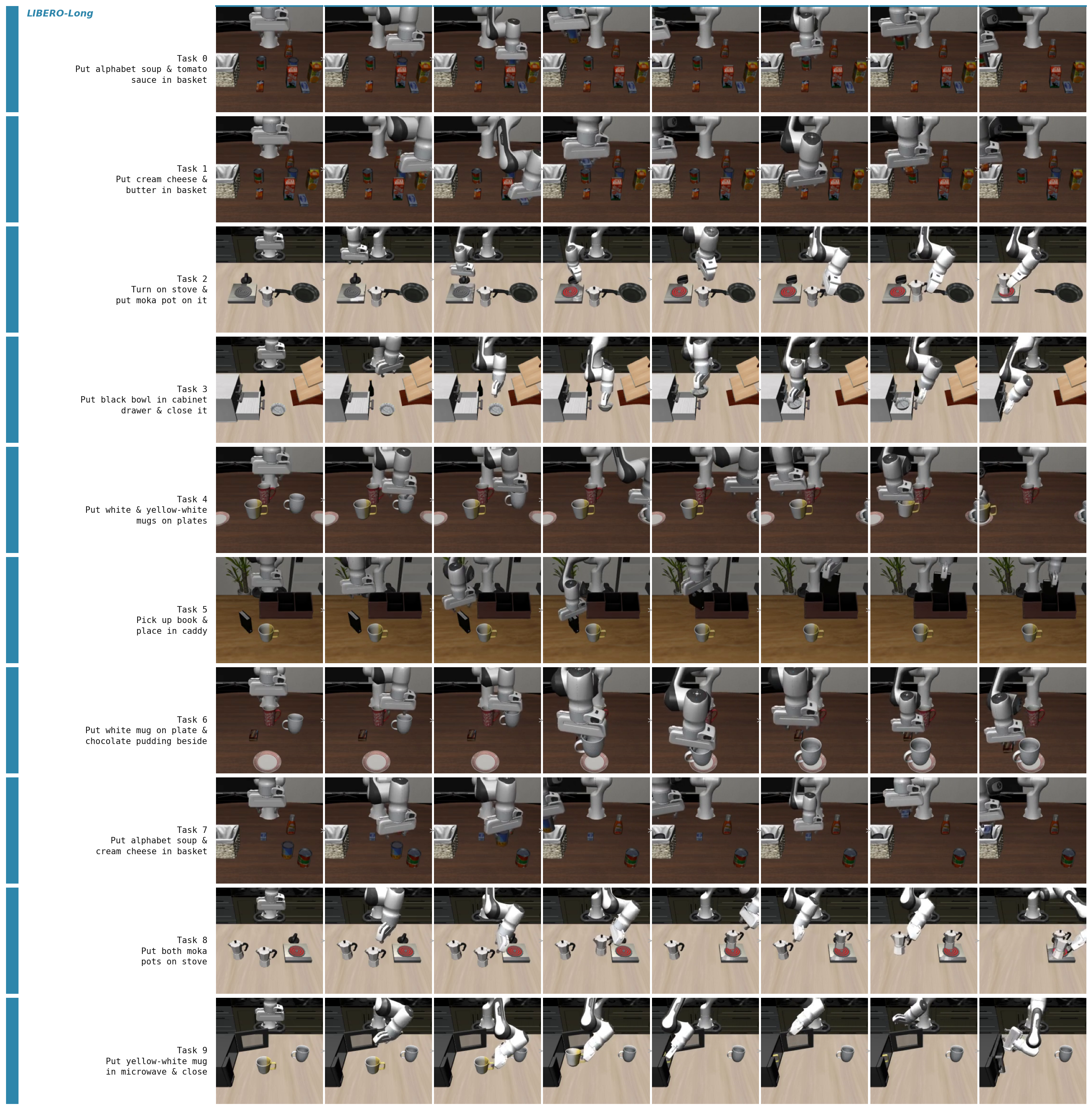}
    \caption{LIBERO-Long tasks overview}
    \label{fig:libero}
\end{figure}

LIBERO-Long contains 10 long-horizon tabletop manipulation tasks (as shown in Fig.~\ref{fig:libero}). 
We collect 50 rollouts for each task, resulting in 500 rollouts per policy. 
The benchmark consists of multi-stage manipulation tasks where the robot must complete multiple object interactions within a single rollout.

We collect rollouts from two policies on LIBERO-Long.
\textbf{OpenVLA} uses the publicly released \texttt{openvla-7b-finetuned-libero-10} checkpoint, a 7B-parameter vision-language-action model fine-tuned on the LIBERO-10 dataset~\citep{kim24openvla}.
\textbf{$\pi_0$-FAST} uses the \texttt{pi0\_fast\_libero} configuration, initialized from the \texttt{pi0\_fast\_base} pretrained model and fully fine-tuned for 30{,}000 steps on the \texttt{physical-intelligence/libero} dataset, with \texttt{replan\_steps=5} (5 simulator steps executed per inference call, action dimension 7, action horizon 10).

Table~\ref{tab:app_libero_openvla} reports the per-task rollout statistics for OpenVLA, and Table~\ref{tab:app_libero_pi0} reports the corresponding statistics for $\pi_0$-FAST.

\begin{table*}[t]
\centering
\small
\begin{tabular}{cp{0.4\textwidth}ccc}
\toprule
Task ID & Task Name & Success Rate & Avg. Policy Calls & Avg. Sim. Steps \\
\midrule
0 & Put both the alphabet soup and the tomato sauce in the basket & 50.0\% (25/50) & 300.5 & 300.5 \\
1 & Put both the cream cheese box and the butter in the basket & 64.0\% (32/50) & 269.5 & 269.5 \\
2 & Turn on the stove and put the moka pot on it & 58.0\% (29/50) & 289.0 & 289.0 \\
3 & Put the black bowl in the bottom drawer of the cabinet and close it & 42.0\% (21/50) & 285.8 & 285.8 \\
4 & Put the white mug on the left plate and the yellow and white mug on the right plate & 38.0\% (19/50) & 251.7 & 251.7 \\
5 & Pick up the book and place it in the back compartment of the caddy & 72.0\% (36/50) & 192.8 & 192.8 \\
6 & Put the white mug on the plate and put the chocolate pudding to the right of the plate & 56.0\% (28/50) & 258.0 & 258.0 \\
7 & Put both the alphabet soup and the cream cheese box in the basket & 64.0\% (32/50) & 291.2 & 291.2 \\
8 & Put both moka pots on the stove & 32.0\% (16/50) & 410.3 & 410.3 \\
9 & Put the yellow and white mug in the microwave and close it & 54.0\% (27/50) & 323.3 & 323.3 \\
\midrule
\multicolumn{2}{c}{Average} & 53.0\% & 287.2 & 287.2 \\
\bottomrule
\end{tabular}
\caption{
Per-task performance and rollout statistics for OpenVLA (\texttt{openvla-7b-finetuned-libero-10}) on LIBERO-Long with \texttt{replan\_steps=1}. Each task is evaluated with 50 rollouts. Since \texttt{replan\_steps=1}, one policy call corresponds to one policy-controlled simulation step.
}
\vspace{0.2cm}
\label{tab:app_libero_openvla}
\end{table*}

\begin{table*}[t]
\centering
\small
\begin{tabular}{cp{0.4\textwidth}ccc}
\toprule
Task ID & Task Name & Success Rate & Avg. Policy Calls & Avg. Sim. Steps \\
\midrule
0 & Put both the alphabet soup and the tomato sauce in the basket & 80.0\% (40/50) & 54.6 & 280 \\
1 & Put both the cream cheese box and the butter in the basket & 100.0\% (50/50) & 51.5 & 265 \\
2 & Turn on the stove and put the moka pot on it & 30.0\% (15/50) & 50.2 & 258 \\
3 & Put the black bowl in the bottom drawer of the cabinet and close it & 42.0\% (21/50) & 47.3 & 243 \\
4 & Put the white mug on the left plate and the yellow and white mug on the right plate & 88.0\% (44/50) & 48.8 & 251 \\
5 & Pick up the book and place it in the back compartment of the caddy & 62.0\% (31/50) & 35.2 & 183 \\
6 & Put the white mug on the plate and put the chocolate pudding to the right of the plate & 80.0\% (40/50) & 48.0 & 247 \\
7 & Put both the alphabet soup and the cream cheese box in the basket & 94.0\% (47/50) & 52.2 & 269 \\
8 & Put both moka pots on the stove & 2.0\% (1/50) & 74.0 & 378 \\
9 & Put the yellow and white mug in the microwave and close it & 34.0\% (17/50) & 51.7 & 266 \\
\midrule
\multicolumn{2}{c}{Average} & 61.2\% & 49.2 & 253 \\
\bottomrule
\end{tabular}
\caption{
Per-task performance and rollout statistics for $\pi_0$-FAST (\texttt{pi0\_fast\_libero}, fine-tuned from \texttt{pi0\_fast\_base}) on LIBERO-Long with \texttt{replan\_steps=5}. Each task is evaluated with 50 rollouts.
}
\label{tab:app_libero_pi0}
\end{table*}

\clearpage
\subsection{ManiSkill-Long}
\label{app:maniskill_details}
\begin{figure}[h]
    \centering
    \includegraphics[width=1\linewidth]{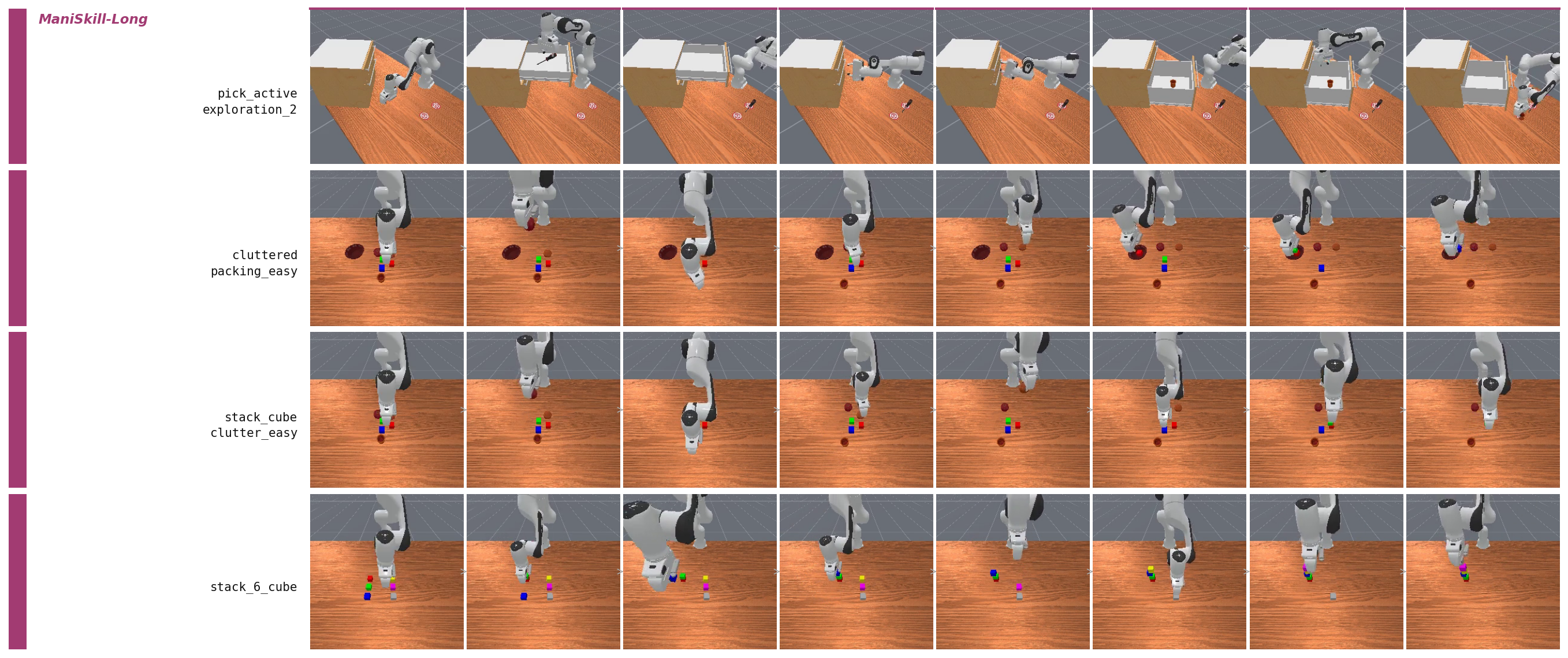}
    \caption{ManiSkill-Long tasks overview}
    \label{fig:maniskill}
\end{figure}

ManiSkill-Long consists of four long-horizon manipulation tasks (as shown in Fig.~\ref{fig:maniskill}) constructed in ManiSkill~\citep{taomaniskill3}. 
These tasks require longer chains of symbolic actions than LIBERO-Long, including exploration, packing, stacking, opening, closing, picking, and placing. 
The tasks are evaluated using the Franka arm embodiment.
\begin{table}[h]
\centering\small
\begin{tabular}{clp{3cm}cccc}
\toprule
Task ID & Short Name & Language Instruction & \#Episodes & SR 
& \shortstack{Avg Policy\\Calls} 
& \shortstack{Avg Sim\\Steps} \\
\midrule
0 & Screwdriver \& Cup & pick up the screwdriver and cup out of the drawer          & 59  & 15\% & 187 & 1807 \\
1 & Cubes into Bowl    & put three cubes into the bowl                               & 100 & 50\% & 152 & 1646 \\
2 & Stack 3 Cubes      & stack 3 cubes together, start with red cube                 & 100 & 50\% & 142 & 1336 \\
3 & Stack 6 Cubes      & stack 6 cubes together, start with red cube                 & 60  & 17\% & 178 & 1122 \\
\bottomrule
\end{tabular}
\vspace{0.1cm}
\caption{Task descriptions and rollout statistics for ManiSkill-Long.
         Exec horizon = 16 sim steps per policy call. 
         Avg sim steps computed over successful rollouts only.}
\label{tab:tasks_maniskill}
\end{table}
We adopt $\pi_0$-FAST with the \texttt{pi0\_maniskill\_rlds\_finetune} checkpoint, initialized from \texttt{pi0\_fast\_base} and LoRA fine-tuned for approximately 100{,}000 steps on the ManiSkill RLDS dataset. Training used a cosine learning rate schedule (peak $10^{-4}$, decay $10^{-5}$, 2{,}000 warmup steps), with action dimension 8 and action horizon 16.
The training demonstrations were generated automatically without human teleoperation: a PDDL-based task planner decomposes each task into a sequence of symbolic actions (pick, place, stack, etc.), which are then executed by MPlib~\citep{taomaniskill3}, ManiSkill's built-in motion planning library that uses the RRTConnect~\cite{kuffner2000rrtconnect} algorithm from OMPL~\citep{sucan2012the-open-motion-planning-library} to compute collision-free joint trajectories.
Table~\ref{tab:tasks_maniskill} summarizes the task-level rollout statistics for $\pi_0$-FAST. In total, we collect 319 valid rollouts across four tasks.

Compared with LIBERO-Long, ManiSkill-Long requires longer execution horizons. 
Successful $\pi_0$-FAST rollouts require 93 policy calls and 1,484 simulation control steps on average.


\clearpage

\subsection{BEHAVIOR-1K}
\label{app:b1k_details}

\begin{figure}[!htbp]
    \centering
    \includegraphics[width=1\linewidth]{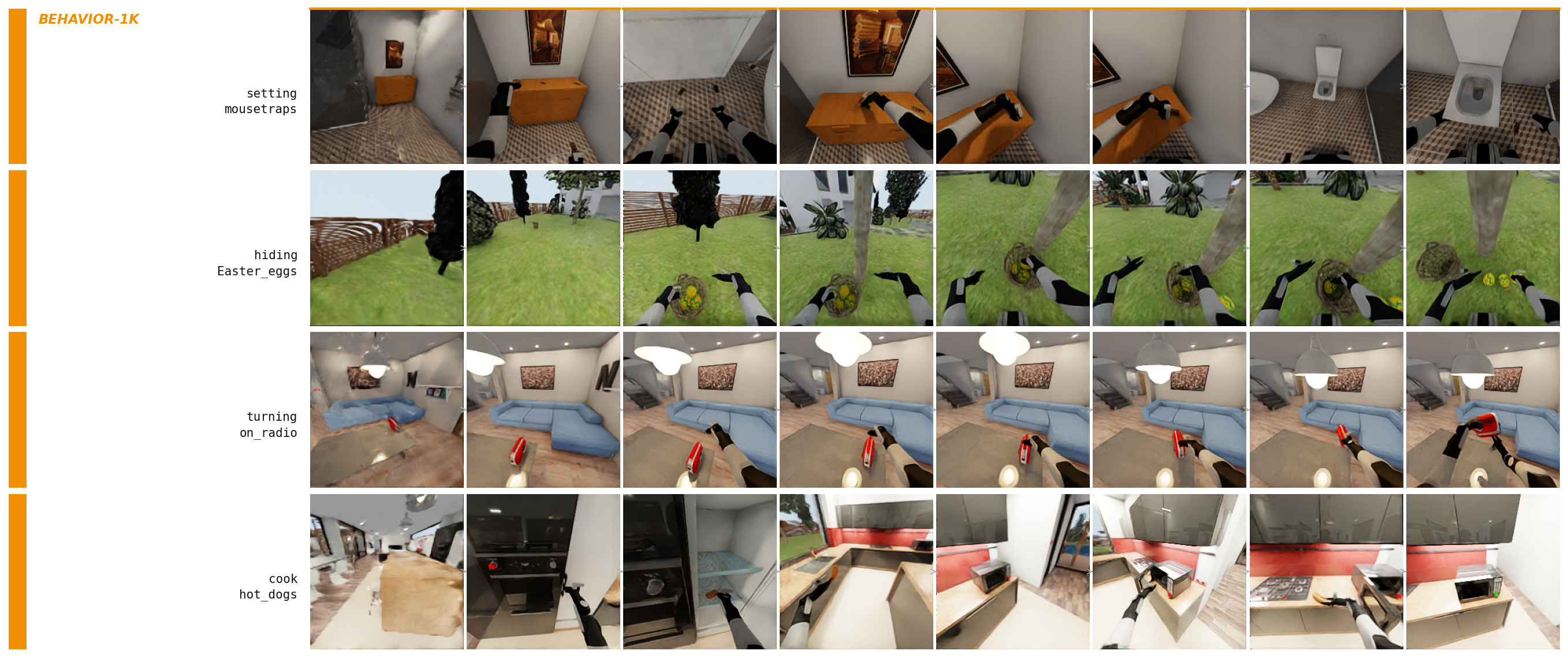}
    \caption{Behavior-1k tasks overview}
    \label{fig:b1k}
\end{figure}

BEHAVIOR-1K evaluates long-horizon mobile manipulation in large-scale household environments. 
We select four tasks (as shown in Fig.~\ref{fig:b1k}) from the BEHAVIOR-1K challenge long-horizon benchmark. 
Unlike LIBERO-Long and ManiSkill-Long, which use a fixed-base Franka arm, BEHAVIOR-1K uses the R1Pro mobile manipulator and requires both navigation and manipulation.(See Table~\ref{tab:tasks_b1k} for details)

We use a revised version of $\pi_{0.5}$, initialized from the \texttt{pi0.5\_base} pretrained checkpoint and fine-tuned on the BEHAVIOR-1K challenge demonstration dataset~\citep{li2024behavior1k}. This model was the 1st-place solution in the 2025 BEHAVIOR-1K Challenge (26\% overall q-score)~\citep{larchenko2025behavior}. Fine-tuning used 200{,}000 steps with delta actions, action horizon 30, action dimension 23 (zero-padded to 32 internally), and 50 trainable task embeddings replacing text conditioning. The four evaluated tasks used the same checkpoint.

We collect 100 rollouts per task, resulting in 400 rollouts in total. Table~\ref{tab:tasks_b1k} reports the task-level rollout statistics. Rollouts were collected targeting 50 successes and 50 failures per task for \ours{} training; the reported success rate reflects the policy's natural success rate observed during collection. The average successful rollout requires 427.4 policy calls and 8,557 simulation steps (1 policy call = 20 simulator steps). The longest task, \texttt{setting\_mousetraps}, requires 13,657 simulation steps on average.

\begin{table}[!htbp]
\centering\small
\begin{tabular}{clp{3.4cm}cccc}
\toprule
Task ID & Short Name & Language Instruction & \#Episodes & SR 
& \shortstack{Avg Policy\\Calls} 
& \shortstack{Avg Sim\\Steps} \\
\midrule
3  & Setting Mousetraps & Take the four mousetraps from the cabinet in the bathroom and place them on the bathroom floor, at least two next to the same sink. & 100 & 50\% & 849 & 13657 \\
4  & Hiding Easter Eggs & Take the three Easter eggs out of the wicker basket on the lawn and place them next to a single tree, none left in the basket.          & 100 & 50\% & 595 &  8540 \\
10 & Turning On Radio   & Turn on the radio receiver that's on the table in the living room.                                                                    & 100 & 50\% & 167 &  2375 \\
47 & Cook Hot Dogs      & Take the two hot dogs out of the refrigerator in the kitchen and cook them in the microwave until both are cooked.                        & 100 & 50\% & 695 &  9654 \\
\bottomrule
\end{tabular}
\vspace{0.1cm}
\caption{Task descriptions and rollout statistics for BEHAVIOR-1K.
         Exec horizon = 20 sim steps per policy call. All 4 tasks are seen (3-fold cross-validation).
         Language instructions from the official BEHAVIOR-1K challenge~\citep{li2024behavior1k}.
         }
\label{tab:tasks_b1k}
\end{table}

\subsection{Real-World Benchmarks}

\begin{figure}
    \centering
    \includegraphics[width=1\linewidth]{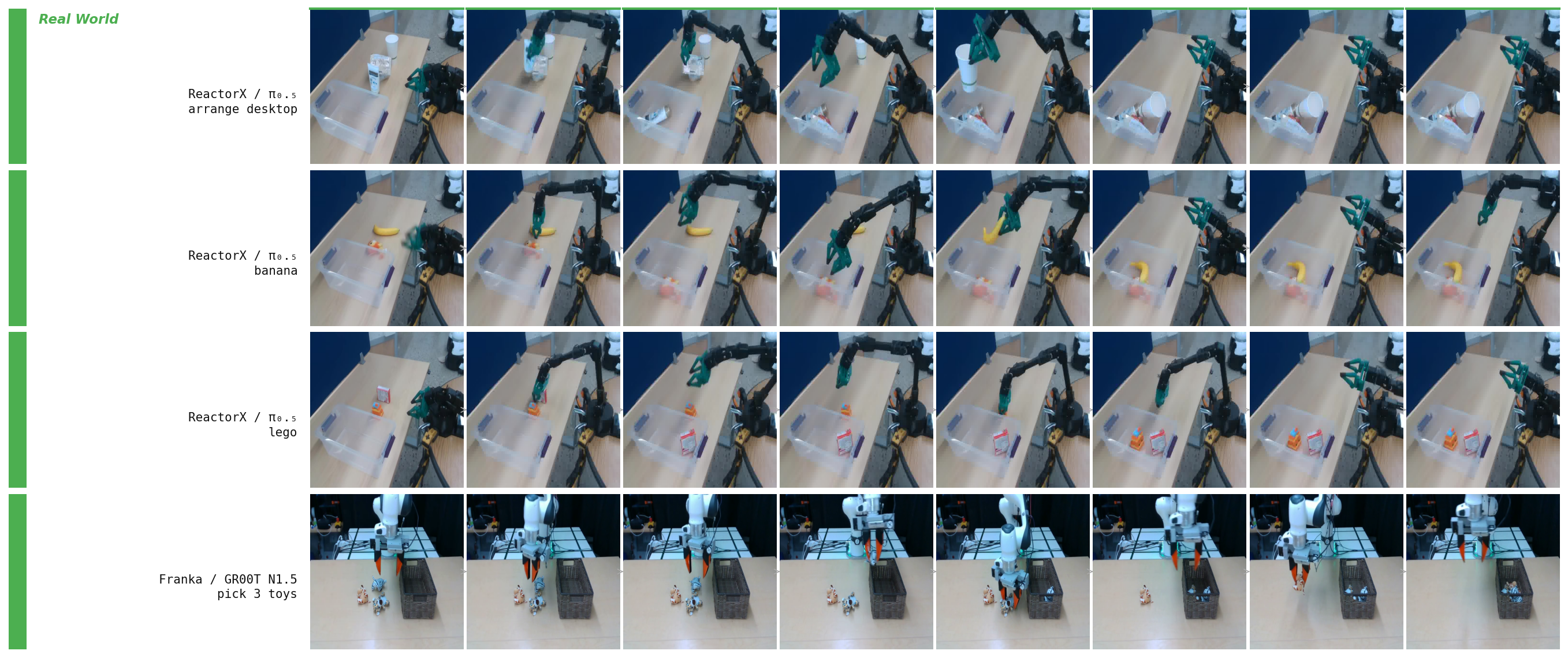}
    \caption{Real-world experiment task overview.}
    \label{fig:real_world}
\end{figure}

We evaluate on four real-robot platforms spanning three policies on a ReactorX arm and one policy on a Franka arm (as shown in Fig.~\ref{fig:real_world}).
Table~\ref{tab:tasks_realworld} summarizes task descriptions, episode
counts, success rates, and rollout lengths.

\paragraph{ReactorX / ACT.}
We collect 40 episodes per task (banana, lego, arrange) using an
ACT policy~\citep{zhao2023learning} trained for 50k gradient steps.
Each episode consists of 12 policy calls at an execution horizon of
100 steps per call ($\sim$1150 total executed steps).
Success rates range from 8\% (lego, a precision-intensive task) to
50\% (arrange).

\paragraph{ReactorX / $\pi$0.5 and SmolVLA.}
We collect 40 episodes per task for both $\pi_{0.5}$~\citep{black2025pi05} and
SmolVLA~\citep{shukor2025smolvla}, each with 14 policy calls per episode
($\sim$1190 total executed steps).
SmolVLA achieves notably higher success on arrange (65\%) compared to
$\pi_{0.5}$ (22\%), reflecting differences in policy capability on
the more structured placement task.

\paragraph{Franka / GR00T N1.5.}
We collect 44 episodes of the ``pick 3 toys'' task using
GR00T N1.5~\citep{gr00tn1_2025} on a Franka arm, with an average of 38
policy calls and an exec horizon of 45 steps per call
($\sim$1700 total executed steps), achieving 48\% success.

\begin{table*}[!htbp]
\centering\small
\setlength{\tabcolsep}{4pt}
\begin{tabular}{llp{3cm}cccc}
\toprule
Robot / Policy & Task & Language Instruction & \#Ep. & \makecell{SR} & \makecell{Avg Policy\\Calls} & \makecell{Avg Sim\\Steps} \\
\midrule

\multirow{3}{*}{ReactorX / ACT}
  & Banana  & pick up banana and lion toy into basket             & 40 & 48\% & 12 & 1141 \\
  & Arrange & put skin cream, praline, and bottle into the basket & 40 & 50\%  & 12 & 1178 \\
  & Lego    & pick up lego jiggler box                            & 40 &  8\%   & 12 & 1146 \\
\midrule
\multirow{3}{*}{ReactorX / $\pi_{0.5}$}
  & Banana  & pick up banana and lion toy into basket             & 40 & 18\%   & 14 & 1188 \\
  & Lego    & pick up lego block and cleaner into basket          & 40 & 18\%   & 14 & 1190 \\
  & Arrange & put skin cream, praline, and bottle into the basket & 40 & 22\%   & 14 & 1190 \\
\midrule
\multirow{3}{*}{ReactorX / SmolVLA}
  & Banana  & pick up banana and lion toy into basket             & 40 & 28\% & 14 & 1190 \\
  & Lego    & pick up lego block and cleaner into basket          & 40 & 30\%  & 14 & 1192 \\
  & Arrange & put skin cream, praline, and bottle into the basket & 40 & 65\%  & 14 & 1192 \\
\midrule
Franka / GR00T N1.5
  & Pick 3 Toys & pick up 3 toys                                 & 44 & 48\% (21/44) & 38 & ${\sim}$1727 \\
\bottomrule
\end{tabular}

\caption{
Task descriptions and rollout statistics for real-world benchmarks.
Each episode runs until task completion or a fixed time limit.
For Franka, average steps are estimated as average policy calls $\times$ execution horizon (45 steps/call).
}
\vspace{0.1cm}

\label{tab:tasks_realworld}
\end{table*}

\section{Ablation Studies}
\label{app:ablations}

This section studies which components of \ours{} are responsible for performance.

\subsection{World-Model Backbone}
\label{app:backbone_ablation}

\begin{table}[t]
\centering
\small
\setlength{\tabcolsep}{6pt}
\begin{tabular}{lccc}
\toprule
Representation model 
& MLP 
& LSTM 
& Transformer \\
\midrule
Cosmos-Predict2.5-2B robot-AC 
& $0.85 \pm 0.02$ 
& $0.85 \pm 0.01$ 
& $0.84 \pm 0.02$ \\
V-JEPA 2-AC 
& \textbf{$0.88 \pm 0.01$} 
& \textbf{$0.86 \pm 0.02$} 
& \textbf{$0.89 \pm 0.02$} \\
\bottomrule
\end{tabular}

\caption{
World-model backbone comparison on LIBERO-Long. We compare ROC-AUC using action-conditioned representations from Cosmos-Predict2.5-2B AC and V-JEPA 2-AC across MLP, LSTM, and Transformer detectors. Values are reported as mean $\pm$ standard deviation across folds. Best results are shown in bold.
}
\label{tab:cosmos_vjepa_libero}
\end{table}

\paragraph{Cosmos-Predict2.5-2B finetuning and feature extraction.}
We finetune the pretrained \texttt{nvidia/Cosmos-Predict2.5-2B} robot action-conditioned
checkpoint~\citep{agarwal2025cosmos} on LIBERO-Long using LoRA (rank~32, $\alpha$~=~32) applied to the
attention and MLP projection layers of the video DiT
(\texttt{q/k/v/output\_proj}, \texttt{mlp.layer1/2}), yielding approximately
20M trainable parameters out of 2B total. Each fold is trained for 5{,}000 iterations
(batch size 1) at 256$\times$320 resolution.
At inference time, for each policy timestep $t$ we feed the current observation frame
and a look-ahead action chunk of $A{=}12$ future 7-DoF scaled delta-EEF actions to the
model, which predicts $A{+}1{=}13$ total latent frames at spatial resolution
$32{\times}40$ with 16 channels.
The conditioning frame occupies latent index 0; we retain the three predicted
future temporal tokens at indices 1--3
(corresponding to $\lfloor A/4 \rfloor = 3$ future latent frames), yielding a tensor
of shape $(16, 3, 32, 40)$.
The latent feature vector at timestep $t$ is obtained by averaging over the
spatial and temporal dimensions, producing a 16-dimensional representation
that encodes action-conditioned predicted future dynamics.

The results suggest that V-JEPA-style latent prediction provides stronger failure-detection features than diffusion-based video generation on LIBERO-Long. 

A likely explanation is that failure detection does not require pixel-level details which are hard to predict, but requires representations for predictable aspects of a scene, exposing robot-object state and action-conditioned deviations from expected dynamics.

\clearpage
\subsection{Hidden Latents versus Action-Conditioned Predicted Latents}
\label{app:hidden_vs_pred}

The comparison between $z^h$ and $z^p$ tests whether action conditioning improves failure detection. 
Hidden latents $z^h$ primarily summarize the current visual observation, while predicted latents $z^p$ encode the world model's action-conditioned expectation of how the scene should evolve. 
This distinction is important because many robot failures are not visually anomalous in isolation; they are mismatches between the intended action and the observed state transition.

\begin{table*}[!htbp]
\centering
\small
\setlength{\tabcolsep}{5pt}
\begin{tabular}{lccc ccc}
\toprule
Benchmark
& \multicolumn{3}{c}{Hidden latent $z^{\mathrm{hidden}}$}
& \multicolumn{3}{c}{Predicted latent $z^{\mathrm{pred}}$} \\
\cmidrule(lr){2-4}\cmidrule(lr){5-7}
& MLP & LSTM & Transformer
& MLP & LSTM & Transformer \\
\midrule
LIBERO-Long 
& 0.77 $\pm$ 0.02 & 0.83 $\pm$ 0.00 & 0.85 $\pm$ 0.02
& 0.88 $\pm$ 0.01 & 0.86 $\pm$ 0.02 & 0.89 $\pm$ 0.02 \\
ManiSkill-Long 
& 0.74 $\pm$ 0.02 & 0.78 $\pm$ 0.02 &  0.81 $\pm$ 0.02 
& 0.70 $\pm$ 0.03 & 0.76 $\pm$ 0.00 &  0.83 $\pm$ 0.02 \\

\bottomrule
\end{tabular}
\caption{
Latent-feature ablation for V-JEPA 2-AC. We compare ROC-AUC using hidden latents before the action-conditioned layer and predicted latents after the action-conditioned layer across benchmarks. Values are reported as mean $\pm$ standard deviation across folds.
}
\label{tab:latent_hidden_vs_pred}
\end{table*}

\clearpage
\section{Qualitative Results}
\label{app:qualitative}

We present qualitative examples of \ours{} (Transformer, predicted states) across all four benchmarks.
Each figure shows ten uniformly sampled frames from the rollout (bottom strip) alongside the full failure score curve $s_t$ and the calibrated functional conformal prediction threshold $\delta_t$ (top panel).
Frame borders are coloured \textcolor{ForestGreen}{\textbf{green}} when $s_t < \delta_t$ (safe) and \textcolor{red}{\textbf{red}} once the alarm latches at the first crossing $t^* = \inf\{t \colon s_t \geq \delta_t\}$.
For each benchmark we show one \emph{true negative} (successful episode that remains below the threshold throughout) and one \emph{true positive} (failing episode where \ours{} raises an alarm before task termination).
The conformal miscoverage levels $\alpha$ used are $0.02$ (LIBERO-Long), $0.02$ (ManiSkill-Long), $0.20$ (BEHAVIOR-1K), and $0.10$ (real-world), matching the values reported in the main results table.


\begin{figure}[h]
    \centering
    \includegraphics[width=\linewidth]{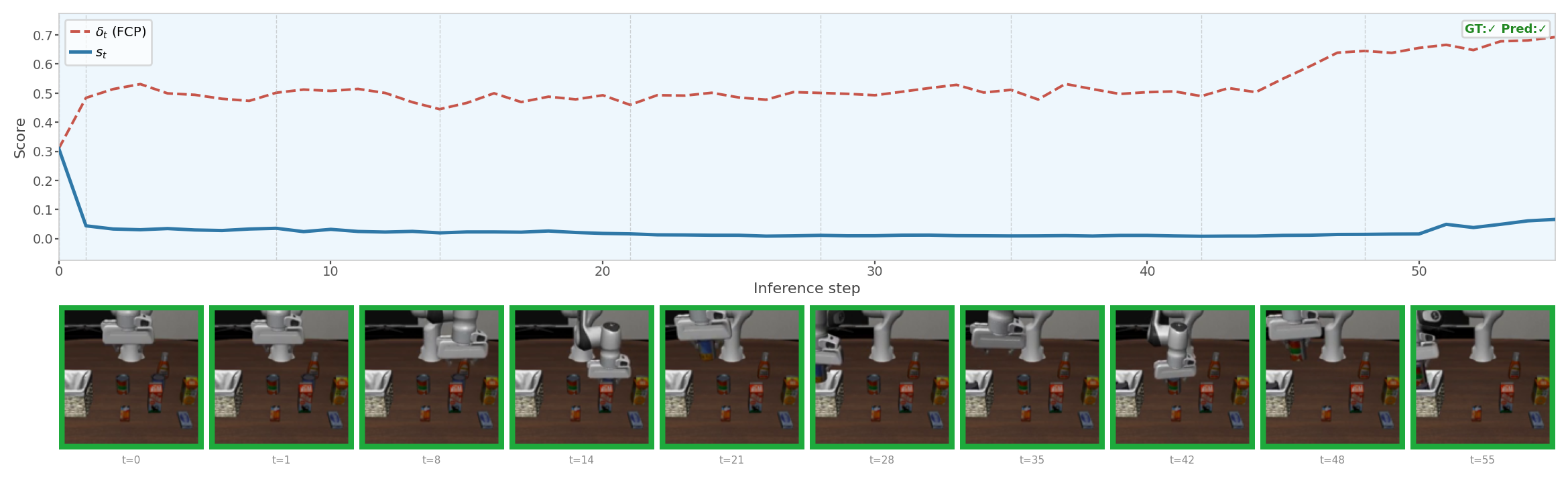}
    \caption{
        \textbf{LIBERO-Long (True Negative)} ($\alpha{=}0.02$, Task~0).
        \emph{``Put both the alphabet soup and the tomato sauce in the basket.''}
        The failure score $s_t$ (blue) remains below the FCP threshold $\delta_t$ (red dashed)
        throughout all inference steps; no alarm is raised and all frame borders are green.
    }
    \label{fig:libero_tn}
\end{figure}

\begin{figure}[h]
    \centering
    \includegraphics[width=\linewidth]{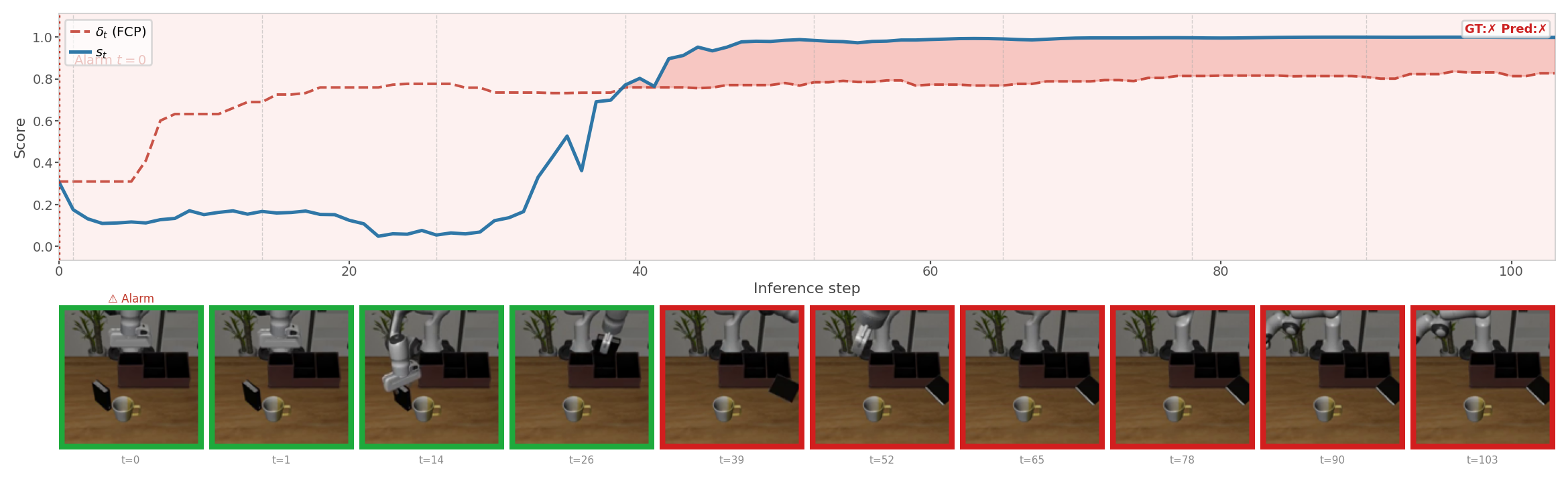}
    \caption{
        \textbf{LIBERO-Long (True Positive)} ($\alpha{=}0.02$, Task~5).
        \emph{``Pick up the book and place it in the back compartment of the caddy.''}
        \ours{} raises an alarm before episode termination as the
        action-conditioned world model's predicted states increasingly diverge from observed states.
        The robot failed the task because it dropped the book during the middle of execution.
    }
    \label{fig:libero_tp}
\end{figure}


\begin{figure}[h]
    \centering
    \includegraphics[width=\linewidth]{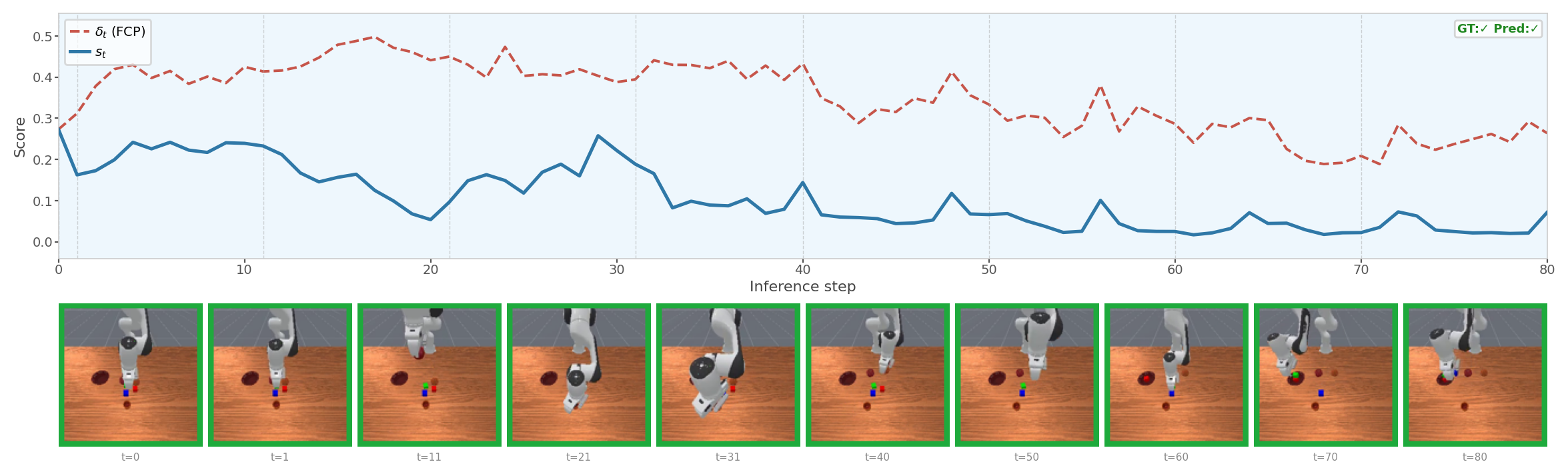}
    \caption{
        \textbf{ManiSkill-Long (True Negative)} ($\alpha{=}0.02$, Task~2: \emph{Cubes into Bowl}).
        \emph{``Put three cubes into the
bowl.''}
    }
    \label{fig:maniskill_tn}
\end{figure}

\begin{figure}[h]
    \centering
    \includegraphics[width=\linewidth]{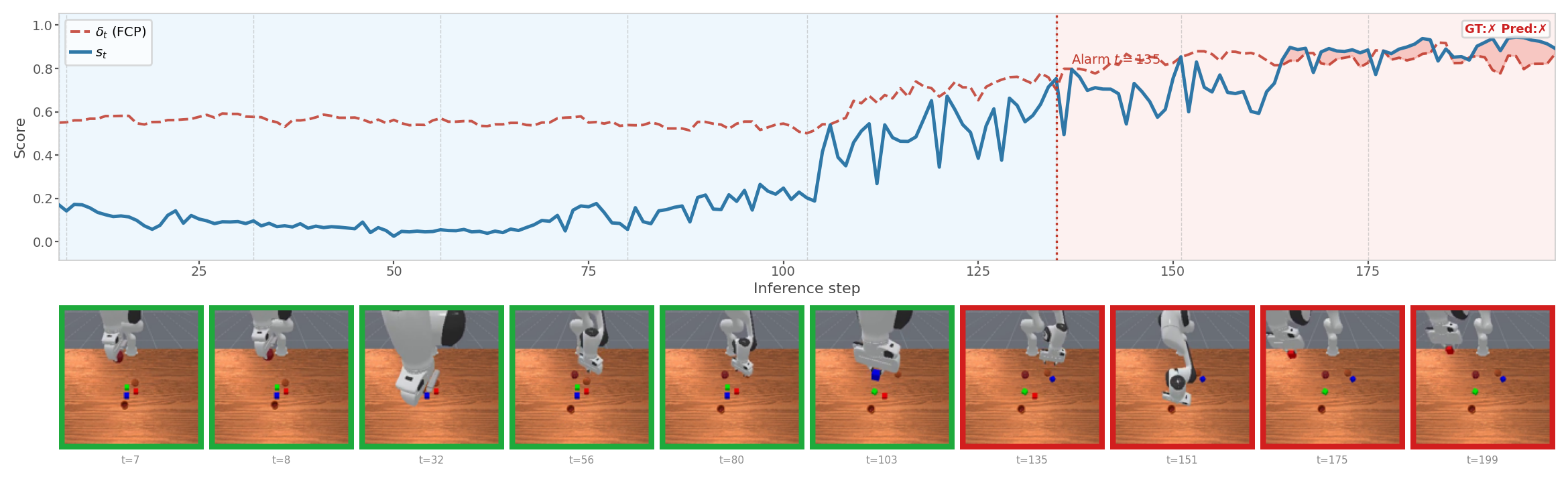}
    \caption{
        \textbf{ManiSkill-Long (True Positive)} ($\alpha{=}0.02$, Task~3: \emph{Stack 3 Cubes }).
        \emph{``Stack 3 cubes together, starting with the red cube.''}
        The robot failed to stack the red cube on the blue cube, leading to the final failure.
    }
    \label{fig:maniskill_tp}
\end{figure}


\begin{figure}[h]
    \centering
    \includegraphics[width=\linewidth]{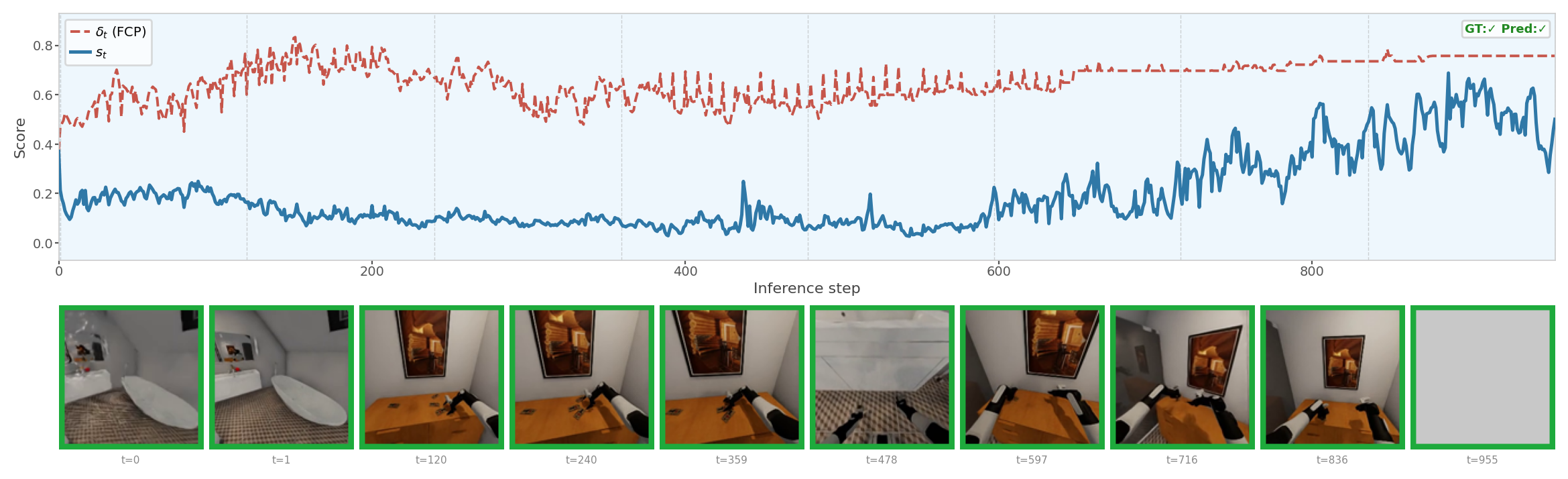}
    \caption{
        \textbf{BEHAVIOR-1K (True Negative)} ($\alpha{=}0.20$, Task~3: \emph{Setting Mousetraps}).
        \emph{``Take four mousetraps from the bathroom cabinet and place at least two next to the
        same sink.''}
    }
    \label{fig:b1k_tn}
\end{figure}

\begin{figure}[h]
    \centering
    \includegraphics[width=\linewidth]{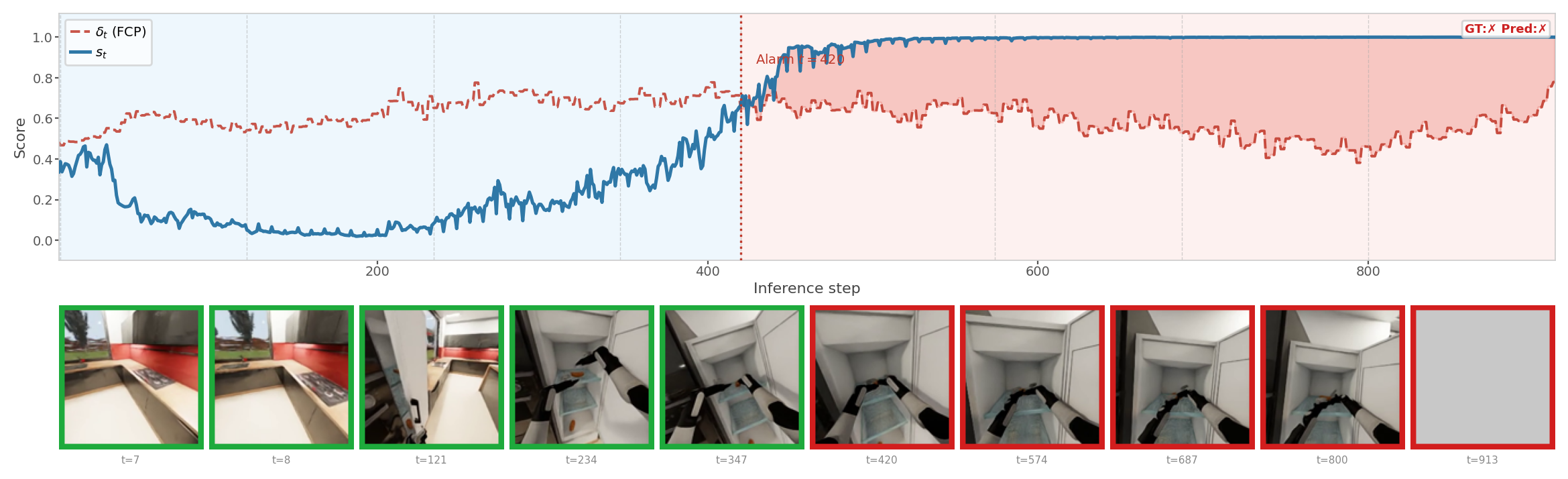}
    \caption{
        \textbf{BEHAVIOR-1K (True Positive)} ($\alpha{=}0.20$, Task~47: \emph{Cook Hot Dogs}).
        \emph{``Take two hot dogs from the refrigerator and cook them in the microwave.''}
        The robot fails during this task because it did not grasp the first hot dog.
    }
    \label{fig:b1k_tp}
\end{figure}


\begin{figure}[h]
    \centering
    \includegraphics[width=\linewidth]{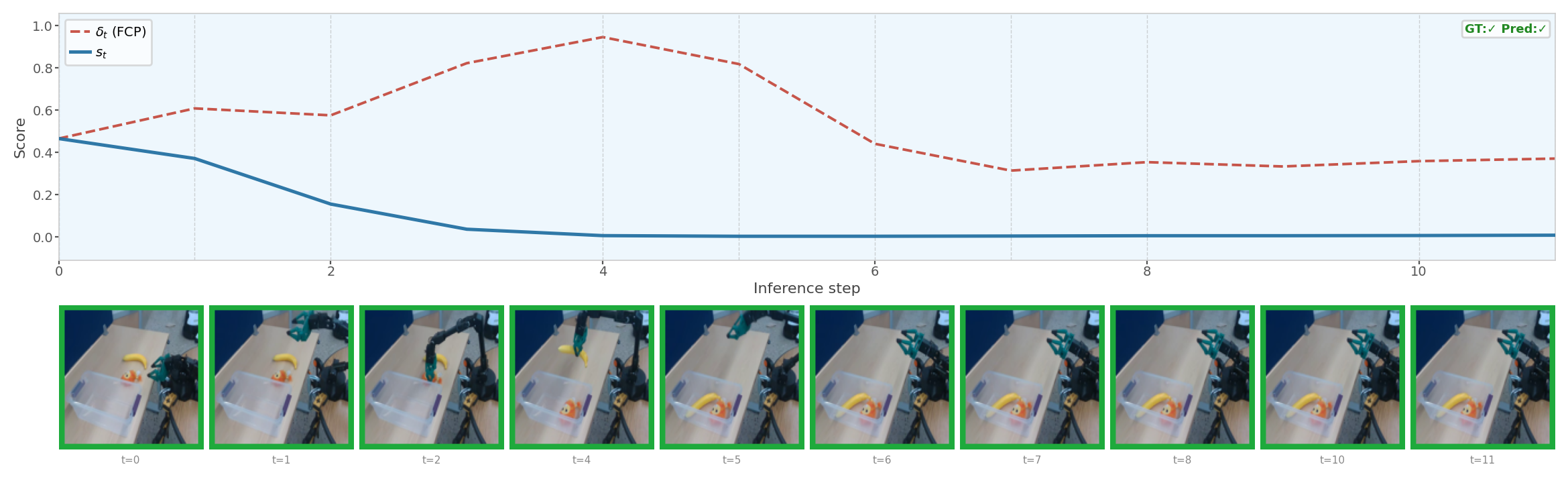}
    \caption{
        \textbf{Real-world (ReactorX\,/\,ACT) (True Negative)} ($\alpha{=}0.10$, Pick Banana and toy lion task).
        \emph{``Pick up banana and lion toy into basket.''}
        No false alarm is raised, showing \ours{} does not penalize successful executions.
    }
    \label{fig:realworld_tn}
\end{figure}

\begin{figure}[h]
    \centering
    \includegraphics[width=\linewidth]{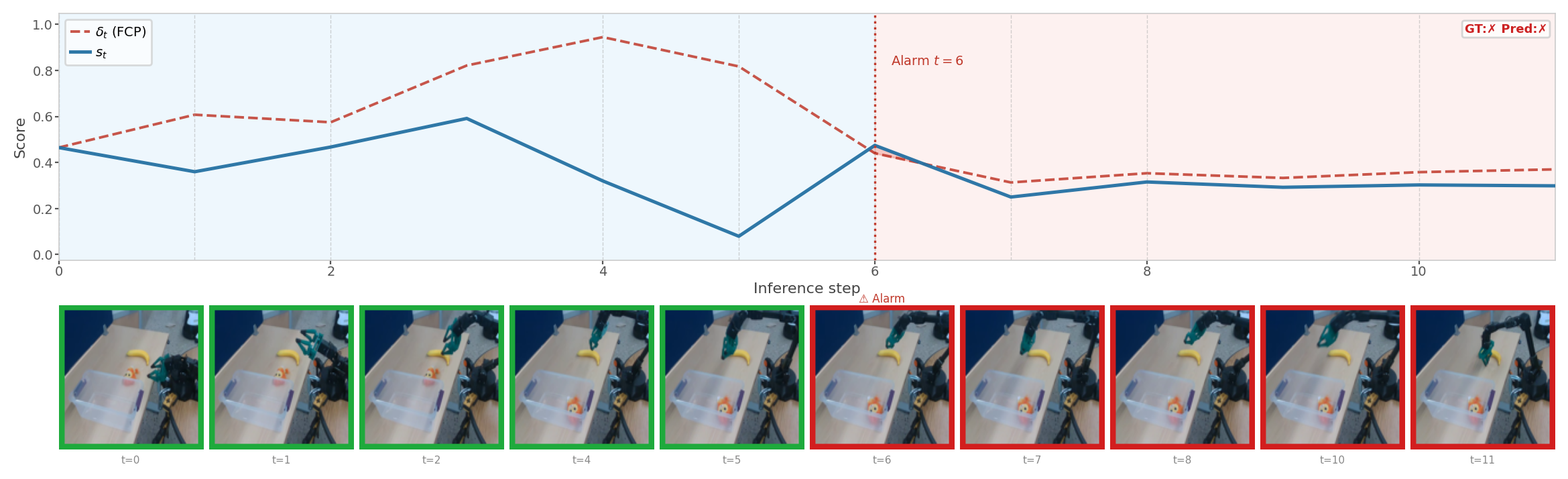}
    \caption{
        \textbf{Real-world (ReactorX\,/\,ACT) (True Positive)} ($\alpha{=}0.10$, Pick Banana and toy lion task).
        \emph{``Pick up banana and lion toy into basket.''}
        A failing real-robot episode from the same task.
        The robot failed to pick up the banana, leading to final task failure.
    }
    \label{fig:realworld_tp}
\end{figure}

\clearpage
\section{Runtime and Deployment}
\label{app:runtime_deployment}

\ours{} is intended for runtime monitoring, so deployment cost is an important practical consideration.
All measurements are conducted on a single NVIDIA H200 GPU using CUDA timing events, averaged over 100 forward passes after 10 warm-up iterations.

\subsection*{Inference Latency}

Table~\ref{tab:latency} reports per-call inference latency for each stage of the \ours{} pipeline and for the baselines.
\ours{} is invoked once per action-chunk boundary, so the relevant budget is one replan interval rather than one control step.

\begin{table}[h]
\centering
\caption{
    Inference latency measured on a single NVIDIA H200 GPU (mean $\pm$ std over 100 runs).
    \ours{} is composed of a frozen world-model backbone (V-JEPA 2-AC) and a lightweight failure detector head.
    SAFE baselines operate on features already produced by the policy backbone and therefore incur only the head cost.
    Latency is measured per replan step (every 16 control steps for $\pi_0$-FAST).
}
\label{tab:latency}
\resizebox{\linewidth}{!}{%
\begin{tabular}{llrr}
\toprule
\textbf{Method} & \textbf{Component} & \textbf{Params} & \textbf{Latency (ms)} \\
\midrule
\multirow{5}{*}{\ours{}}
  & World-model encoder (ViT-G/16, 8 frames)   & 1{,}012\,M & $122.54 \pm 0.08$ \\
  & Action-conditioned predictor               & 305\,M     & $60.19 \pm 0.07$  \\
  & \quad Subtotal: feature extraction         & 1{,}317\,M & $182.73 \pm 0.11$ \\
\cmidrule{2-4}
  & Failure detector: MLP                      & 0.4\,M     & $0.08 \pm 0.00$   \\
  & Failure detector: LSTM                     & 3.4\,M     & $0.16 \pm 0.00$   \\
  & Failure detector: Transformer              & 2.0\,M     & $0.91 \pm 0.02$   \\
\cmidrule{2-4}
  & \textbf{Total (MLP head)}                  & 1{,}317\,M & $\mathbf{182.81 \pm 0.00}$ \\
  & \textbf{Total (LSTM head)}                 & 1{,}317\,M & $\mathbf{182.89 \pm 0.00}$ \\
  & \textbf{Total (Transformer head)}          & 1{,}317\,M & $\mathbf{183.64 \pm 0.02}$ \\
\midrule
\multirow{4}{*}{Baselines}
  & SAFE-MLP         & 0.4\,M     & $0.08 \pm 0.00$   \\
  & SAFE-LSTM        & 3.4\,M     & $0.16 \pm 0.00$   \\
  & RND         & 211\,M     & $9.68 \pm 0.04$   \\
  &FAIL-Detect  & 124\,M     & $6.28 \pm 0.02$   \\
\bottomrule
\end{tabular}%
}
\end{table}

\subsection*{Cost Decomposition and Deployment Implications}

The world-model backbone (V-JEPA 2-AC) dominates the total monitoring cost, accounting for over 99\% of \ours{}'s inference time.
The failure detector head itself is negligible regardless of architecture: the MLP, LSTM, and Transformer heads add less than 1\,ms atop the 182.73\,ms backbone cost.
In comparison, SAFE baselines incur no backbone overhead, their policy encoder is already executed during normal action inference, so their effective marginal cost reduces to that of the head alone ($<$0.2\,ms for MLP/LSTM, 6--10\,ms for the diffusion-based RND and FAIL-Detect variants).

Despite this gap, \ours{}'s absolute latency of ${\approx}183$\,ms remains well within the deployment budget of one replan interval.
For $\pi_0$-FAST, which executes a 16-step action chunk before replanning, \ours{} is queried at the chunk boundary rather than at every control step, providing a temporal window of 16 control steps in which to complete inference.

\end{document}